\documentclass[lettersize,journal]{IEEEtran}
\usepackage{url}
\usepackage{textcomp, gensymb}
\usepackage{verbatim, comment} % For verbatim text and comments
\usepackage{blindtext} % For filler text

\usepackage{amsmath, amsfonts, amssymb, bm}
\usepackage{mdwmath, mdwtab, eqparbox} % For math and tables

\usepackage{array}         % For advanced table layouts
\usepackage{booktabs}      % For professional table rules (\toprule, \midrule, \bottomrule)

\usepackage{caption}       % For caption formatting (e.g., \captionsetup)

\usepackage{array, multirow, booktabs, makecell} % For table formatting

\usepackage{graphicx, epstopdf, ifpdf}
\usepackage[pdftex, colorlinks=true, allcolors=blue]{hyperref}

\usepackage{stfloats, fixltx2e}
\usepackage[caption=false,font=normalsize,labelfont=sf,textfont=sf]{subfig}
\usepackage{caption} % For captions

\usepackage{algorithm, algorithmicx, algpseudocode}

\usepackage[normalem]{ulem} % For underlining
\usepackage{adjustbox} % For adjustments
\usepackage{lettrine} % For drop caps

\begin{document}
	
\title{
Design and Benchmarking of A Multi-Modality Sensor for Robotic Manipulation with GAN-Based Cross-Modality Interpretation
}
% <-this % stops a space
\author{
Dandan Zhang*, Wen Fan*, Jialin Lin, Haoran Li, Qingzheng Cong, Weiru Liu, Nathan F. Lepora, Shan Luo
\thanks{D. Zhang, W. Fan, J. Lin, Q. Cong are with the Department of Bioengineering, and Imperial-X Initiative, Imperial College London; H. Li, W. Liu, N. Lepora are with the School of Engineering Mathematics and Technology, and Bristol Robotics Laboratory, University of Bristol. S. Luo is with the Department of Engineering, King's College London.  
}
}
\maketitle

\begin{abstract}
In this paper, we present the design and benchmark of an innovative sensor, ViTacTip, which fulfills the demand for advanced multi-modal sensing in a compact design. A notable feature of ViTacTip is its transparent skin, which incorporates a `see-through-skin' mechanism. This mechanism aims at capturing detailed object features upon contact, significantly improving both vision-based and proximity perception capabilities. In parallel, the biomimetic tips embedded in the sensor's skin are designed to amplify contact details, thus substantially augmenting tactile and derived force perception abilities. To demonstrate the multi-modal capabilities of ViTacTip, we developed a multi-task learning model that enables simultaneous recognition of hardness, material, and textures. 
To assess the functionality and validate the versatility of ViTacTip, we conducted extensive benchmarking experiments, including object recognition, contact point detection, pose regression, and grating identification. To facilitate seamless switching between various sensing modalities, we employed a Generative Adversarial Network (GAN)-based approach. This method enhances the applicability of the ViTacTip sensor across diverse environments by enabling cross-modality interpretation.
\end{abstract}

\section{Introduction}
Given  the complexity of the real-world environment,  a single sensing modality or method for data acquisition may prove inadequate for achieving a comprehensive perception. 
This has led to a notable demand in research aiming at integrating multi-modality sensing in robotic systems for advanced manipulation tasks. These approaches encompass many strategies, such as combining features acquired from visual and tactile sensors within a shared latent space to augment robot perception \cite{luo2018vitac}. Alternatively, some methods focus on mapping representations from one sensory modality to another~\cite{lee2019touching} or leveraging information from one modality to guide exploration in another~\cite{jiang2022shall}. In these studies, visual cameras are employed to capture visual data, while tactile sensors are utilized to gather contact information.
Most prior research has focused on integrating various sensors into robotic systems and fusing the resulting data for analysis. In this work, we aim to develop a unified multi-modality sensor that endows robotic systems with versatile perception capabilities, including vision, tactile, proximity, and force sensing. The proposed sensing system significantly reduces hardware and computing requirements in compact robotic systems while significantly enhancing integration capabilities.

\begin{figure}[t]
	\centering
\captionsetup{font=footnotesize,labelsep=period}
\includegraphics[width = 1\hsize]{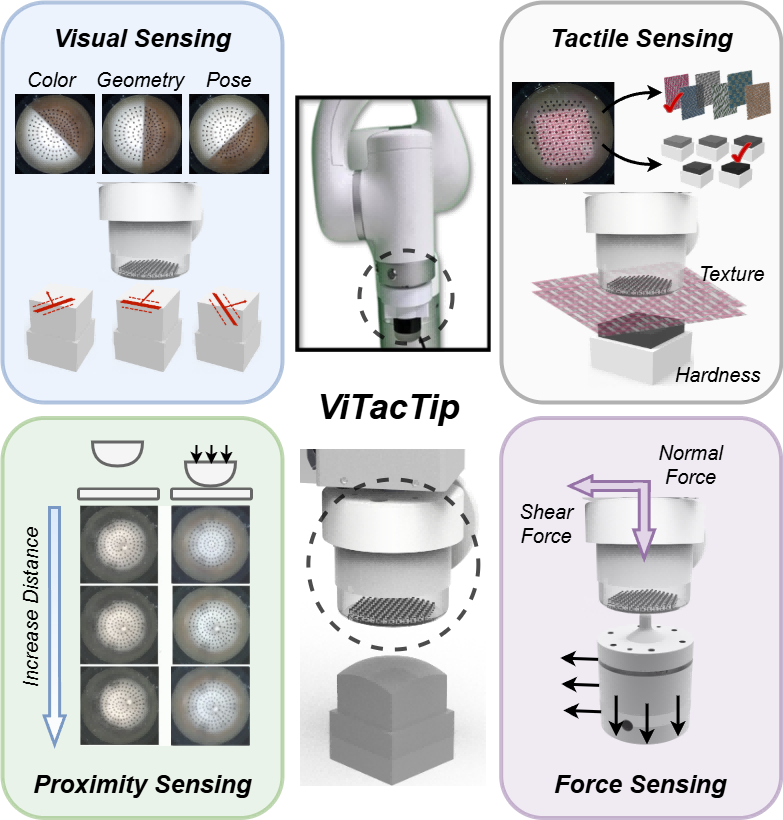}
	\caption{Overview of the four perception capabilities of ViTacTip, including two principle modalities: visual and tactile sensing, as well as two derived modalities: proximity and force sensing.}
 \vspace{-0.2cm}
	\label{pose1}
\end{figure}

One of the critical challenges in multi-modal sensor fusion is minimizing cross-sensitivity, which occurs when the input from one modality interferes with the output of another. This cross-sensitivity complicates the decoupling of data from different modalities, making it difficult to extract information specific to a particular modality for downstream tasks. To address these challenges, recent research has focused on strategies such as employing distinct signal patterns, implementing control mechanisms in composite materials, and optimizing the morphology of active materials \cite{yang2022multimodal}.
Developing sensors that utilize materials with frequency-dependent characteristics, such as ionic/proton-based materials, has been explored for their potential to decouple stimuli \cite{yang2022multimodal}. Combining innovative structures with functional materials like piezoresistive electrodes in capacitive sensors has enabled multi-modal sensing from a single unit \cite{szczuko2022evaluation}. However, the aforementioned approaches necessitate specialized hardware development, potentially complicating the manufacturing process and increasing hardware expenses significantly. 

The recent development of Generative Adversarial Networks (GANs)-based method \cite{goodfellow2020generative, lin2023attention, saxena2021generative} can be used to transfer sensing modality from one domain to another.  Therefore, we aim to explore GANs-based approaches to enable seamless transitions among different data visualization modes, allowing the sensor to dynamically adapt to varied perceptual demands in diverse operational contexts.

In this paper, we introduce a novel perception system that combines multiple sensory modes (visual, tactile, proximity, and force) within a compact design, as shown in Fig. \ref{pose1}. Its unique combination of the see-through-skin mechanism, biomimetic pin design, and GANs-based modality transition establishes ViTacTip as an advantageous multi-modality sensor. We aim to expand the field of multi-modality sensing by exemplifying its significant role in robotic perception technologies that can be used for complicated tasks in diverse environments. The sensor is designed as an integrated unit, capable of interpreting and responding to various types of stimuli. 

The  \textbf{main contributions} are listed as follows:
\begin{itemize}
\item The innovative development of the ViTacTip sensor, a multi-modal fusion device capable of capturing both tactile information and visual details.

\item The implementation of a GAN-based methodology to enable seamless switching between visual and tactile sensing modalities.

\item The comparative studies of ViTacTip against the open-source TacTip sensor \cite{lepora2021soft} and our in-house developed ViTac sensor.

\end{itemize}

In this work, we further elaborate on the multi-modality sensing capabilities of ViTacTip by providing a detailed analysis of its visual and tactile sensing, along with two derived modalities (proximity and force perception). Additionally, we significantly expand the experimental evaluation of the overall methodology. The additional \textbf{contributions} presented in this paper are as follows:

\begin{itemize}
\item We detail the multi-modal perception capabilities of ViTacTip and present hardware benchmarking experiments covering contact point detection, pose regression, and grating identification.

\item We introduce a hierarchical multi-task learning framework to demonstrate ViTacTip's advanced multi-modal perception capabilities through hardness, material, and texture recognition.

\item We present a comprehensive analysis of the GAN-based cross-modality interpretation framework, including both qualitative and quantitative evaluations of the usability and robustness of the Light Removal GAN (LR-GAN) and Marker Removal GAN (MR-GAN) across diverse tasks and complex environments.
\end{itemize}

The remainder of this paper is structured as follows: Section \ref{related} presents an overview of the related work in this field. Section \ref{Design-Fabrication} details the design and fabrication processes of ViTacTip. The multi-modality sensing capabilities of ViTacTip are thoroughly explained in Section \ref{Multi-Modality}. Section \ref{hardware} focuses on hardware benchmarking, including comparative studies with baseline sensors.   Subsequently, Section \ref{GAN} introduces the development and application of LR-GAN and MR-GAN for effective sensing modality transitions. Discussions on the findings and potential avenues for future research are presented in Section \ref{Discuss-Future}. Finally, Section \ref{Conclusions} concludes the paper by summarizing the key insights and contributions.

\section{Related Work}
\label{related}
\subsection{Vision-Based Tactile Sensors}
Most of the existing vision-based tactile sensors (VBTSs) are primarily categorized into several distinct types. The first type encompasses coated-type sensors, which are characterized by their reflective layers, with the GelSight family being a prominent example.  GelSight \cite{yuan2017gelsight} features a unique design with a thin reflective layer atop a clear elastomer layer, which is further supported by a flat acrylic plate. This configuration, illuminated parallel to the surface, facilitates the detection of deformations using photometric stereo methods. 
Subsequent advancements have further refined this technology \cite{8593661}. For instance, GelTip \cite{9340881} has enabled curved sensing surfaces with a single camera. Additionally, OmniTact \cite{9196712} employs five cameras to augment multi-directional sensing capabilities.

The second category encompasses purely marker-based sensors, which are available in various forms. For example, the GelForce sensor \cite{kamiyama2005vision} incorporates markers suspended within an elastomer.  Such movement of the pins introduces an amplification effect in the shear motion of these markers, significantly improving the tactile sensor's responsiveness to both shear and normal deformations on its detection surface. 
Moreover, the TacTip skin is made by 3D-printing which leads to a wide range of different customized designs~\cite{lepora2021soft}.
In addition, DTac-type sensors \cite{lin2023dtact}, such as 9DTac \cite{lin20239dtact} and C-Sight \cite{fan2024design}, employ a combination of translucent gel and an opaque layer. Variations in pixel darkness within the images allow for intensity-to-depth regression and can even facilitate force and torque estimation.

A fundamental principle of VBTSs is the transduction of tactile information into visual data, using the sensor's skin as the medium \cite{shimonomura2019tactile, fan2024crystaltac, fan2024magictac}. Tactile information is transformed into visual features, such as gel deformation, shadows, or marker movements, which are captured by internal cameras. This conversion enables the processing of tactile features leveraging the strengths of state-of-the-art deep learning-based vision techniques \cite{fan2022graph, abderrahmane2019deep, fan2023tac}.

However, as single-modality tactile sensors, both GelSight and TacTip have a significant limitation: their inability to incorporate visual perception. In GelSight, the reflective coating inhibits light transmission, while in TacTip, the black skin blocks external light. As a result, these designs exclude critical information about the touched object, such as fine surface textures and pre-contact distance perception.
This exclusion of visual data can hinder a comprehensive understanding of tactile perception processes, especially in tasks involving interactions with real-world physical environments \cite{lee2019touching}. These limitations motivate the development of a new sensor that integrates both visual and tactile modalities to enhance perception capabilities for robotic applications.

\subsection{Multi-Modality Sensors}
Multi-modality data, which combines both visual and tactile information from the same object, has been shown to significantly improve performance in texture recognition tasks \cite{cao2023multimodal}. Therefore, the development of a well-designed multi-modality sensor capable of simultaneously capturing visual and tactile data during interactions is of considerable importance.

FingerVision \cite{yamaguchi2016combining}, a representative multi-modality sensor, is fabricated entirely from transparent silicone. It can acquire visual information directly through the sensor's skin. However, this design is limited by the absence of an internal light source, making its reliability highly susceptible to variations in environmental lighting conditions. Moreover, the integration of visual and tactile perception modalities in this sensor is tightly coupled, potentially affecting its robustness and adaptability to diverse tasks.

SpecTac \cite{9812348} introduces an innovative approach to multi-modal sensing by integrating ultraviolet (UV) LEDs and randomly distributed UV fluorescent markers. When illuminated by the UV LEDs, these markers become visible, allowing for clear differentiation and tracking against the background. SpecTac enables seamless switching between visual and tactile sensing modes by controlling the activation of the UV LEDs, which toggles the visibility of the markers. However, its performance is constrained under low-light conditions.

TIRgel \cite{10224334} employs a lens-shaped elastomer and an adjustable-focus camera to enable switching between visual and tactile sensing modes. It leverages Total Internal Reflection (TIR) within the elastomer for tactile imaging. However, TIRgel features a relatively flat surface and lacks force-sensing capabilities, which limits its suitability for dynamic manipulation tasks.

Most of the aforementioned work utilizes hardware approaches to switch between different modalities. 
To the best of our knowledge, the software-based approach (such as the adoption of GANs) for modality switching in multi-modality sensors remains rarely explored. 
Our primary objective is to create a multi-modal sensor that includes vision, tactile, proximity, and force sensing functions. Additionally, we plan to gather extensive information and improve the sensor's robustness across different environmental settings. To accomplish this target, we plan to utilize GAN-based models. These models are intended to ease the transition between sensing modes and boost the sensor's overall performance.

\begin{figure*}[!htbp]
	\centering
\captionsetup{font=footnotesize,labelsep=period}
\includegraphics[width = 1\hsize]{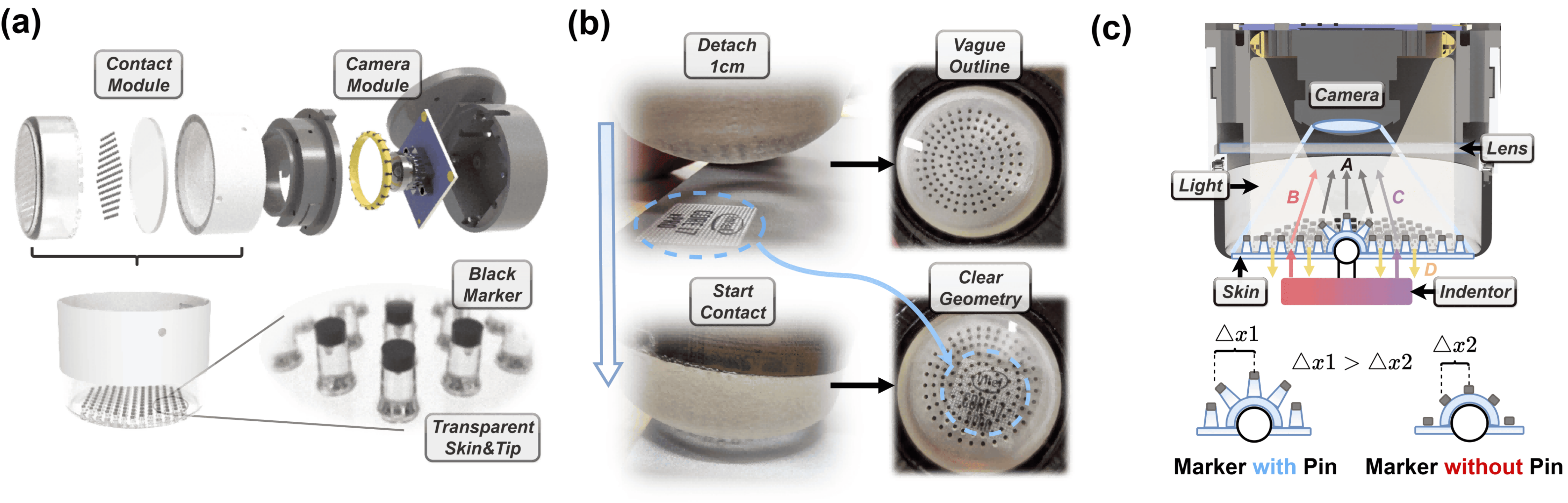}
	\caption{(a) Architecture of the ViTacTip Sensor: An exploded view illustrating the sub-components, a side view of the assembled ViTacTip, and a detailed illustration of the sensor's tips and markers. (b) Demonstration of ViTacTip's proximity and vision perception capabilities.
(c) Sketch of ViTacTip’s working principles: A: Projections of tactile deformations mapped by the displacement (${\Delta}x$) of black markers on the sensor tips. B/C: Visual feature projections captured through the sensor's transparent skin. D: Internal lighting passes through the skin, illuminating nearby objects. The sketch also compares the sensor's performance with and without a pin-like marker design. }
 % \vspace{-0.1cm}
	\label{Fig1}
\end{figure*}

\section{Design and Fabrication}
\label{Design-Fabrication}
\subsection{Design Considerations}
Our goal in designing the ViTacTip sensor is to combine multi-modal sensing's versatility with the design inspired by the human fingertip \cite{ward2018tactip}. We aim to replicate the human skin's layered functionality, including the dermal and subcutaneous layers \cite{lepora2021soft}. 
The sensor features a slim, flexible, and transparent rubber-like surface, designed with pin-like formations to resemble the touch-sensitive properties of human skin.  Inside, it houses a transparent, highly flexible polymer that simulates the mechanical characteristics of the skin's deeper layers, which enhances the sensor's tactile response. An internal camera tracks markers on the internal structure of the rubber skin, allowing for precise, remote, and sensitive detection of surface deformations \cite{ward2018tactip}. A ring-shaped light source is used to project uniform illumination downward, highlighting black markers on the tips of internal pins within the camera's view. The overview of the sensor's key components with an exploded view is shown in Fig. \ref{Fig1} (a).

The transparent outer layer of the ViTacTip sensor serves a dual purpose. First, it facilitates proximity sensing, enabling the sensor to estimate the distance to objects approaching its surface without requiring direct contact, as illustrated in Fig. \ref{Fig1} (b). Second, it allows for the visual observation of object features. Upon contact, the soft, flexible skin conforms to the object's shape, delivering detailed tactile information and supporting force measurements. This functionality is further enhanced by an internal supporting gel, which adjusts the orientation of embedded pins to amplify the richness of touch information.

The sketch of ViTacTip’s working principles can be found in  Fig. \ref{Fig1} (c), including the illustrations of i) the projections of tactile deformations mapped by the pin-like black markers, ii ) visual feature projections captured through the sensor's transparent skin, iii) internal lighting passes through the skin. The sketch also compares the sensor's performance with and without the pin-like marker design. It demonstrates that the displacement of the markers, denoted as ${\Delta}x$, is significantly greater than that of the traditional dot-like marker when the sensor interacts with the same indenter.

\subsection{Fabrication of ViTacTip}
The detailed manufacturing process of the ViTacTip sensor is comprehensively illustrated in Fig. \ref{design and fabrication}. The fabrication process is adapted from the open-source designs for the TacTip sensor\footnote{https://softroboticstoolkit.com/tactip}. However, the materials used for the fabrication of TacTip and ViTacTip are different. The black skin of the TacTip blocks the external light source for visual perception. TacTip primarily focuses on tactile sensing without integrated vision capabilities. In contrast, ViTacTip integrates vision modalities seamlessly with tactile sensing via adopting a transparent skin, which enhances the sensor's ability to capture and interpret detailed multi-modal data. More specifically, ViTacTip uses materials with high light transmission properties that enhance the perception of visual features.
As shown in Fig. \ref{design and fabrication} (a), the overall ViTacTip sensor consists of eight sub-assemblies, which can be categorized into three major modules: structure, contact, and perception.

\subsubsection{Structure Module}

The structure module of the sensor system, which includes the camera base and housing part, serves as both a foundational and connective component in the sensor assembly. Fabricated using the Stratasys F370 printer, known for its precision and quality, the module is made from ABS (Acrylonitrile Butadiene Styrene), a thermoplastic polymer valued for its strength and durability.
The high tensile strength of ABS ensures the module's resilience under physical stress, while its excellent impact resistance and toughness safeguard sensitive internal components, such as the cameras. Furthermore, ABS represents an economical manufacturing option, since it offers a balance between cost-effectiveness and high-quality production.

\subsubsection{Contact Module}
\begin{figure*}[!htbp]
\captionsetup{font=footnotesize,labelsep=period}
		\centering
		\begin{tabular}[b]{c}
\hspace{0cm}\includegraphics[width=1\textwidth,trim={0 0 0 0},clip]{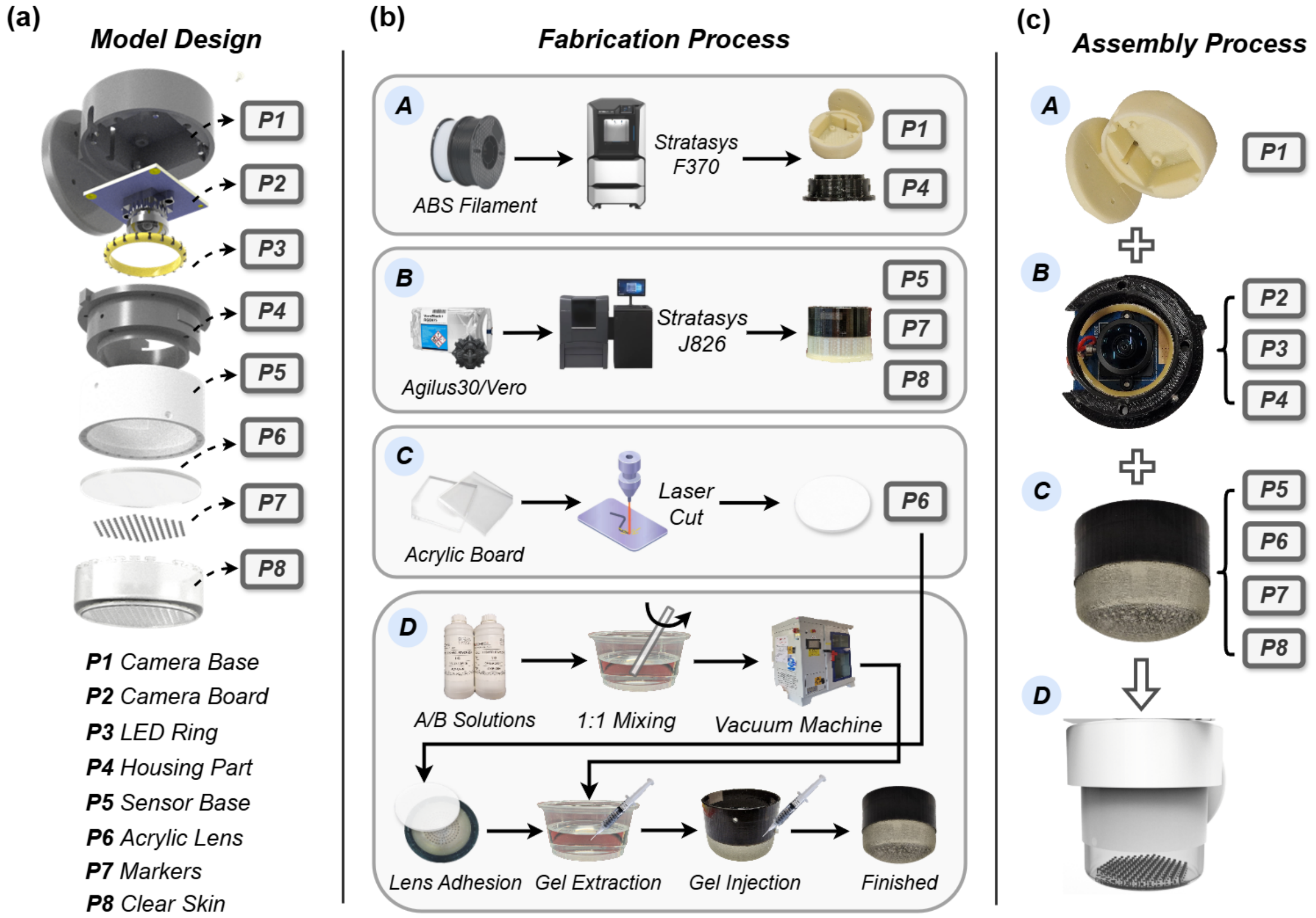} \\
		\end{tabular}
             \caption{The design and fabrication schematic of the ViTacTip:
(a) Entire Model Design: Exploded view of subcomponents.
(b) Fabrication Process: The mounting bases and outer skin are 3D printed. An acrylic lens is glued to the base, and the gel is prepared and injected \cite{lepora2022digitac, lin2022tactile}.
(c) Assembly Process: After fabricating the ViTacTip outer skin, it is assembled with the camera unit, illumination unit, and mounting base to construct the complete system.}
		\label{design and fabrication}%
% \vspace{-0.1cm}
\end{figure*}

The fabrication of the contact module involves a three-step process. As depicted in Fig. \ref{design and fabrication} (b), the assembly of the transparent skin, featuring pin-shaped markers on its inner surface, alongside a rigid base, is accomplished using the multi-material Stratasys J826 3D printer. Specifically, the skin and pins are constructed from Agilus30 Clear, a material known for its flexibility and high light transmission properties. 
It has translucent properties, which makes it suitable for applications where the perception of visual features of external objects is important. 
Agilus30 Clear has a Shore A hardness of 30-35\footnote{https://www.javelin-tech.com/3d/stratasys-materials/agilus-30/}, which offers a blend of flexibility and firmness ideal for conforming to various shapes while maintaining structural integrity. Its polymerized density ranges from 1.14 to 1.15 $g/cm^3$, indicating high durability and stability. With a tensile strength of 2.4-3.1 MPa, it resists stretching and pulling forces, which is crucial for enduring mechanical stresses in tactile sensing. These properties make Agilus30 Clear an ideal material for equipping ViTacTip with the ability to accurately detect and adapt to object contours.

The ViTacTip markers and base are fabricated using VeroBlack, which is selected for its rigidity and opacity \footnote{https://www.dinsmoreinc.com/material/verowhite-veroblue-and-veroblack/}. The pins are initially printed with Agilus30 Clear to ensure flexibility and transparency. Using the multi-material capability of the Stratasys J826 3D printer, the tips of the pins are then overprinted with VeroBlack in a seamless single print cycle. This process ensures precise placement of the black markers at the pin ends, resulting in robust, well-defined markers that contrast effectively with the transparent portions.

After the 3D printing process, a post-processing step is necessary to remove the support material. The ViTacTip markers and base incorporate support structures crucial for preserving their integrity and geometry during the printing process. Support removal can be achieved either through chemical dissolution, where a chemical bath selectively dissolves the support material, or via a high-pressure water jet, which removes the support physically without chemical exposure. We primarily use the water-jet method in this paper.

After cleaning up, a 1.5mm thick acrylic board is laser-cut into a suitable shape and fixed on the print part through glue. Then, in the final step, a gel-like soft material is prepared by mixing two solutions, TECHSTL RTA27905A/B, in a 1:1 ratio. Similar to the preparation of silicone, bubble removal through a vacuum machine is required to realize pure transparency. The prepared gel is carefully injected via a syringe into the space between the skin and the lens, with special attention to avoid the introduction of air bubbles. 
After 12 hours of standing in the heating chamber, the internal gel gradually stabilized, and the entire contact module is ready to use.
As shown in Fig. \ref{design and fabrication} (c), the contact module will be assembled with the structure and perception modules to build an entire ViTacTip sensor.

\subsubsection{Perception Module}

The perception module in the sensor system consists of an ELP USBFHD06H-L180 camera\footnote{https://www.elpcctv.com/elp-2mp-full-hd-1080p-uvc-free-driver-low-light-imx322-usb-camera-wide-angle-for-android-p-333.html} and ring-shaped LED lighting, both of which are essential for image data capture. For fair comparative studies during hardware benchmarking experiments, we use the same camera as TacTip  \cite{lepora2021pose}. The camera, equipped with a wide-angle lens, provides a broad field of view for thorough image capture. The flexible ring-shaped LED is selected for housing it to the limited space within the ViTacTip. We use a gelatinous diffusion material to envelop the LED ring. It effectively softens and evens out the light, which helps reduce glare and shadows. Such uniform light distribution is essential for multi-modal sensing, as it ensures consistent illumination necessary for capturing high-quality images, thereby supporting the accurate and reliable perception.

\section{Multi-Modality Sensing}
\label{Multi-Modality}
\subsection{Principles of Multi-Modality Fusion}
The ViTacTip sensor is designed with a transparent skin, allowing LED light to penetrate and illuminate an area that extends more than 20 mm beyond the sensor's surface. This transparent skin enables the sensor to capture visual information about nearby objects, including their contour, color, and position information.  The sensor effectively discerns most features of an object when it is within a 5-10 mm range. 
Generally, as the ViTacTip approaches an object, proximity perception becomes increasingly significant. 
When the sensor is within 4 mm of an object, its visual perception capabilities are significantly enhanced. Upon direct contact, the sensor's functionality expands to include force detection and tactile sensing. This contact activates the sensor's camera, located at its base, to generate a detailed multi-modal fusion image. This image combines visual and tactile data, providing a more comprehensive understanding of the object than single-modal sensors can achieve. Additionally, changes in marker distribution upon contact indicate the  sensor's force and tactile perception features.

In this section, we evaluate the force sensing, proximity sensing, and vision-tactile fusion-based comprehensive recognition capabilities of the ViTacTip. We begin by illustrating the sensor's proximity and force sensing capabilities (see Subsection \ref{proximity-sense} and Subsection \ref{force-modality}, respectively). Next, we explore and validate the sensor's advanced capabilities in capturing a broader range of visual and tactile information through two tasks:
i) shape (object) recognition (see Subsection \ref{object}); and
ii) hardness, material and texture sensing (see Subsection \ref{multi}).

\subsection{Proximity Sensing}
\label{proximity-sense}
\subsubsection{Task Description}

\begin{figure*}[!htbp]
	\centering
\captionsetup{font=footnotesize,labelsep=period}
\includegraphics[width = 1\hsize]{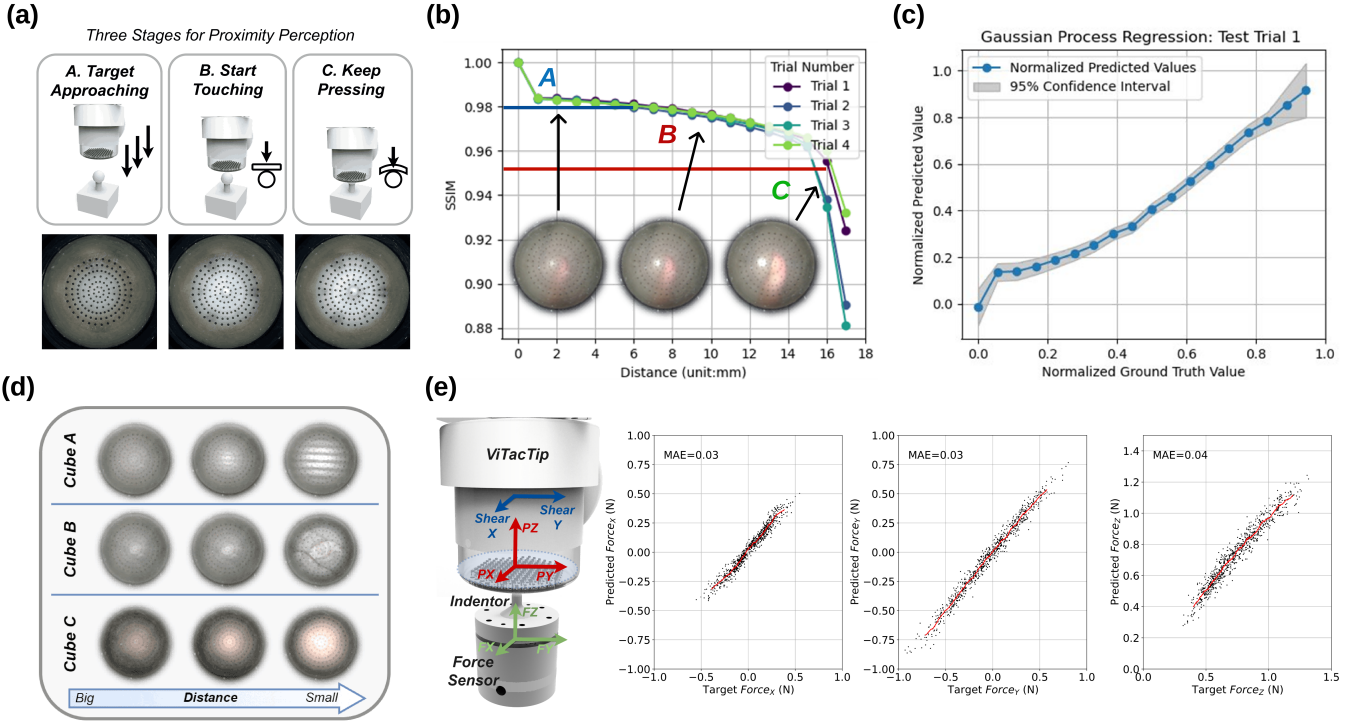}
	\caption{(a) The three stages of the ViTacTip sensing process and the illustration of the proximity perception mechanism.
(b) Curve showing the relationship between distance (0-18 mm) and SSIM values of images obtained from ViTacTip, using a human finger database as an example. Points `A', `B', and `C' correspond to the three stages illustrated in (a). The blue and red lines indicate the thresholds for segmenting the stages between A and B, and B and C, respectively.
(c) GPR-based distance estimation using a database involves approaching human finger with ViTactip.
(d) Examples of images captured during proximity perception between ViTacTip and three cubes with different textures.
(e) The mean average errors (MAE) of ViTacTip in force estimation ($F_x$, $F_y$, $F_z$). Black dots represent the results of the trained model on the test dataset, while the red line represents the smoothed predictions.} 
 \vspace{-0.1cm}
	\label{proximity-update}
\end{figure*}

Proximity sensing is crucial for human-robot interaction and robotic manipulation tasks. Traditional coated-type sensors \cite{yuan2017gelsight} and marker-type sensors \cite{ward2018tactip} (with black skin) lack this capability.
In contrast, the ViTacTip sensor, with its transparent skin, offers a unique proximity-sensing capability. To evaluate this capability, a detailed experiment was conducted by mounting the ViTacTip sensor on the end-effector of a Ufactory 850 robotic arm\footnote{https://www.ufactory.cc/ufactory-850/}. The experiment setup is illustrated in Fig.~\ref{proximity-update} (a).

To evaluate the reliability and variability of distance measurements, we collected data across various scenarios involving interactions with a sharp object of complex shape, a flat desktop, a human finger, and three cubes covered with fabrics featuring different texture patterns, referred to as Cube A, Cube B, and Cube C, respectively. 
The resulting datasets are referred to as the `Sharp Object Dataset', `Human Finger Dataset', `Flat Desktop Dataset', `Cube A/B/C Dataset', respectively.

During experiments, we controlled the ViTacTip to gradually approach the target object and acquire the corresponding perception image in real-time. The sensor was directed towards the object gradually until direct contact was established, which marked a critical point in the experiment. Notably, the robot's movement did not stop at the initial contact; it continued an additional 4mm downward. Each trial of the experiment was divided into distinct phases based on the sensor's distance from the object. In Stage A, when the sensor was positioned within 20-40mm of the object, the initial activation of its proximity detection feature was observed.
As the sensor approached the target, entering Stage B, the sensor started to reflect a more precise visualization of the object. Upon making contact with the object, as depicted in Stage C, a notable change was observed in the sensor's output. This stage was characterized by changes in the distribution of markers, signaling the commencement of tactile and force perception.

We use the Structural Similarity Index Measure (SSIM) \cite{gomes2021generation} to estimate object distances for proximity perception, a choice inspired by previous studies \cite{cao2023toward}. An image captured when the ViTacTip is far from the target object serves as the reference image. The SSIM is then calculated between the real-time image and this reference image.   Fig. \ref{proximity-update} (b) illustrates the relationship between distance and SSIM values of images obtained from ViTacTip, using the human finger database as examples, where points `A', `B', and `C' correspond to the three stages shown in Fig. \ref{proximity-update} (a). The blue and red lines represent the thresholds used to segment the stages between i) Stage A and B, ii) Stage B and C, respectively.

However, the effectiveness of SSIM in proximity sensing is influenced by object characteristics, such as texture, color, and reflectivity \cite{rasti2021image}. For instance, smooth surfaces, such as desks or non-textured walls, often exhibit limited textural variability. This makes distance estimation more challenging compared to textured surfaces, such as a human hand or fabric-covered objects. Therefore, we explore a machine learning-based approach to enhance the robustness for proximity perception capability. More specifically, SSIM is used as a feature for the Gaussian Process Regression (GPR) model-based distance estimation, equipped with a Radial Basis Function (RBF) kernel and supplemented by a constant kernel to account for data variability.

\subsubsection{Results Analysis}

\begin{table}[htbp]
\centering
\captionsetup{font=footnotesize,labelsep=period}
\caption{Summary for Distance Prediction Tests Across Datasets}
\label{tab:combined_ssim_values}
\begin{tabular}{@{}lccccc@{}}
\toprule
\textbf{Dataset} & \textbf{Trial 1} & \textbf{Trial  2} & \textbf{Trial  3} & \textbf{Trial  4} & \textbf{Mean} \\ \midrule
Sharp Object & 0.0009 & 0.0009 & 0.0009 & 0.0012 & 0.0010 \\
Human Finger  &  0.0045 & 0.0007 & 0.0026 & 0.0039 & 0.0029 \\
Flat DeskTop & 0.0074 & 0.0054 & 0.0067 &  0.0045 & 0.0060 \\
 % \bottomrule
 \toprule
 \textbf{Dataset} & \textbf{Trial 1} & \textbf{Trial  2} & \textbf{Trial  3} & \textbf{Trial  4} & \textbf{Mean} \\ \midrule
Cube A  & 0.0016 & 0.0016 & 0.0017 & 0.0015 & 0.0016 \\
Cube B        & 0.0015 & 0.0011 & 0.0021 & 0.0012 & 0.0016 \\ 
Cube C        & 0.0017 & 0.0020 & 0.0022 & 0.0027 & 0.0022 \\
\bottomrule
\end{tabular}
 \vspace{-0.1cm}
\end{table}

The Mean Squared Error (MSE) between two images is calculated to quantify the errors between ground truth distance estimation values and model predictions for proximity perception.
As shown in Table \ref{tab:combined_ssim_values}, in the Sharp Object dataset, MSE values were highly consistent, ranging from 0.0009 to 0.0012, with an average of 0.0010, indicating reliable predictions across trials. In contrast, the Flat Desktop dataset showed greater variation in MSE values, ranging from 0.0045 to 0.0074, with an average of 0.0060.
For the Human Finger dataset, the MSE values remained relatively stable. The average MSE for this dataset was 0.0029, which is slightly higher than that of the Sharp Object dataset but significantly lower than Flat Desktop dataset. This reflects the sensor’s improved ability to capture fine details, compared to simpler desktop scenarios. An example of the comparison between distance estimation using the GPR model and the ground truth data is shown in Fig. \ref{proximity-update} (c).

The GPR model delivered reliable proximity estimation across the Cube datasets. Examples of images captured from proximity perception between ViTacTip and the three different cubes with various textures are shown in Fig. \ref{proximity-update} (d).
For Cube A, MSE values ranged from 0.0015 to 0.0017, with an average of 0.0016, reflecting stable predictions throughout the trials.  Cube B demonstrated similar stability, with MSE values between 0.0011 and 0.0021, also averaging at 0.0016. 
Cube C exhibited slightly greater variation, with MSE values between 0.0017 and 0.0027 and an average of 0.0022. 
These results confirm the model's robustness in estimating proximity for various objects, with minor differences likely due to  texture variations.

The overall performance highlights the ViTacTip sensor's reliability across various proximity scenarios. It performs particularly well in detecting sharp objects, human fingers, and cubes with complex textures (Cube A and B). However, there is some variability in proximity detection for the flat desktop and Cube C, which have simpler textures. This may be due to the lack of distinct features, slightly reducing the sensor's accuracy in SSIM-based proximity perception.
To further enhance the sensor's proximity sensing capabilities, we will incorporate advanced deep learning models capable of adapting to environmental changes and varying object properties in the future.

\subsection{Force Sensing}
\label{force-modality}

\subsubsection{Task Description}

Force sensors are crucial for robotics. However, the high cost of these sensors can often be a barrier for many businesses and individuals who require them.
For example, standard force sensors like the ATI (Axia80-M20)\footnote{https://www.ati-ia.com/index.aspx} or NBIT\footnote{http://www.iibtcn.com} are priced between \$3,000 and \$5,000. Given this high cost, there is an increasing interest in the robotics community to explore tactile sensors as a low-cost alternative to these traditional high-end force sensors.

In our research, we explore the potential of ViTacTip for 3-axis force estimation, while the same method can be extended for 3-axis torque estimation. We conducted an experiment to explore the usage of ViTacTip as a force sensor. We positioned a commercial NBIT force sensor beneath the ViTacTip\footnote{https://www.sensor360.org/brand/7473}, which was mounted on the wrist of a low-cost robotic arm (MG400, Dobot). The robotic arm can be used to control the ViTacTip to approach the targeted force sensor. This setup allowed us to simultaneously gather data from the force sensor and the ViTacTip during their interaction. A dataset of 3,000 images was collected to train a deep neural network model. 
We ensured that the pose values [$X$(mm), $Z$(mm), $\theta$($\degree$)] varied within [-5, 5]  for $X$, [-1, 1] for $Z$, and [-45, 45] for $\theta$, respectively\cite{lepora2022digitac}.

The ViTacTip sensor was designed to prioritize high sensitivity in detecting subtle force variations, a critical requirement in robotic applications for precise control and preventing damage during tasks such as grasping, assembly, and handling. Its small force measuring range ensures operation within its most sensitive regime, enabling accurate detection and measurement of subtle forces. This capability is particularly valuable for safely and efficiently interacting with fragile materials or soft objects \cite{he2020soft}.

\subsubsection{Results Analysis}
The experiment began with a standard press on the sensor and introduced a shear displacement along the horizontal plane. This action was designed to exert a three-dimensional (3D) force on the sensor's surface, simulating real-world scenarios where forces are not limited to a single direction. Additionally, we varied the intensity of external light sources during the experiment to assess the impact of changing ambient light conditions on the sensor’s performance.
Our primary objective was to accurately predict the forces ($Fx$, $Fy$, $Fz$) resulting from normal pressing and the horizontal shearing motions.

As shown in Fig.~\ref{proximity-update} (e), our results demonstrated a remarkable ability to predict 3D forces accurately. Specifically, we observed a horizontal force estimation error between -0.5 and 0.5 N and a normal pressure prediction error of  0.04 N. These results suggest that the ViTacTip not only serves as a feasible alternative to traditional force sensors in terms of cost but also provides reliable and precise force perception. The estimated cost for each ViTacTip sensor is about \$50, including the costs for 3D printing and camera.

\subsection{Multi-Modality Perception for Object Recognition}
\label{object}

\subsubsection{Task Description}
The ability to capture comprehensive object shape information is essential for robots performing a wide range of tasks \cite{luo2017robotic}. Luo et al. established this task as a benchmark for evaluating a tactile sensor's capability to capture shape information, using objects with distinct geometric configurations such as 3D-printed spheres, crescents, triangles, and others \cite{gomes2021generation}.

\begin{figure*}[!htbp]
\captionsetup{font=footnotesize,labelsep=period}
	\centering
\includegraphics[width = 1\hsize]{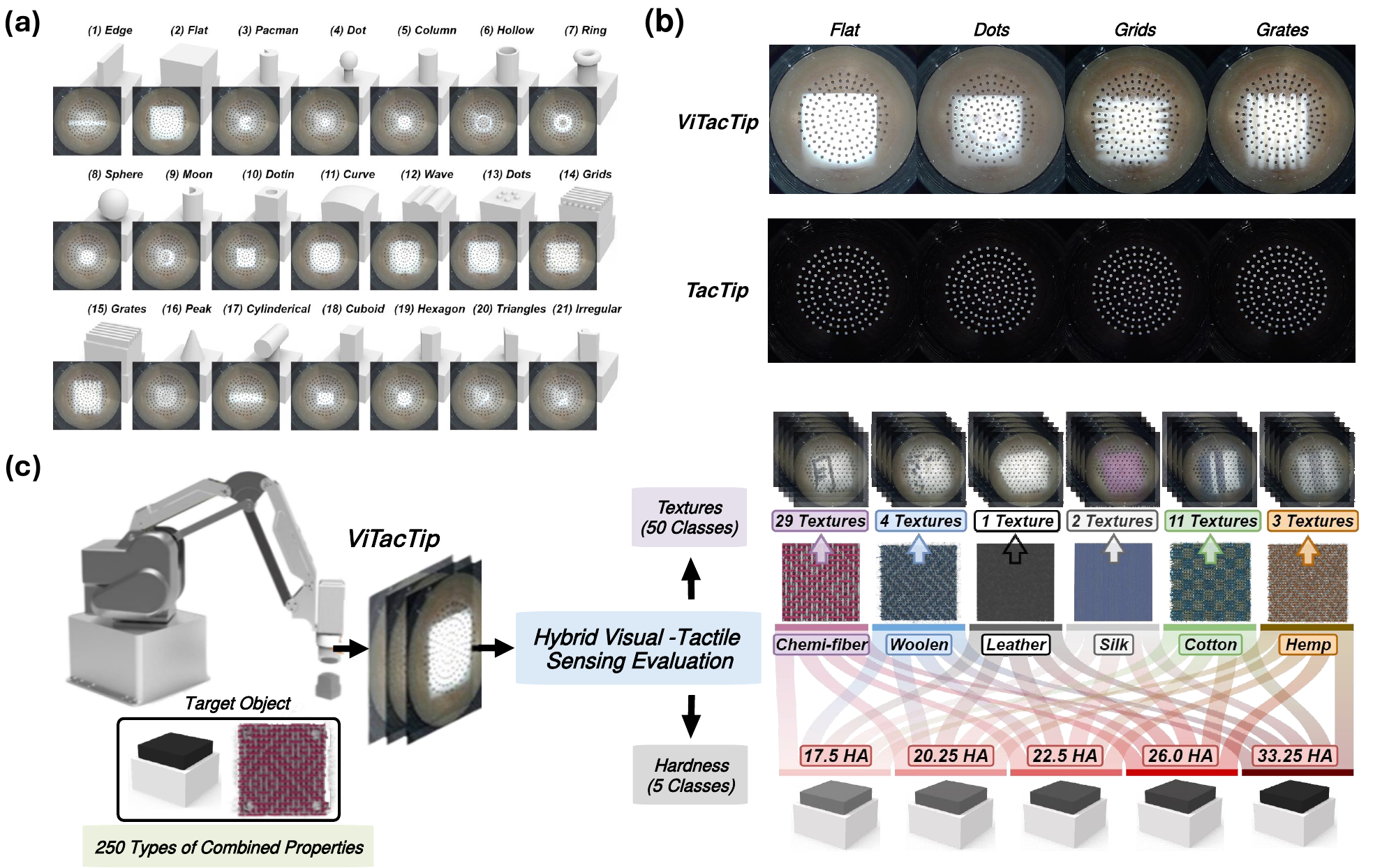}
	\caption{(a) Samples for the object recognition task: 21 objects with different shapes, as detailed in \cite{gomes2021generation}. Real imaging from ViTacTip allows recognition of object localization and contour, with marker distribution adapting to the contact shape. (b) High-resolution perception images from ViTacTip and TacTip obtained by interacting with typical objects, illustrating challenges in shape differentiation using tactile information alone without visual support. (c) Experimental setup: a desktop Dobot robotic arm (MG400) for data collection \cite{lepora2022digitac}, examples of texture samples used in experiments, and a schematic of hybrid visual-tactile sensing evaluation. Covering the elastomer with fabric requires ViTacTip to perceive both the elastomer's hardness and the fabric's texture upon contact.}
 \vspace{-0.1cm}
	\label{hybrid tactile sensing}
\end{figure*}

The ViTacTip sensor is well-suited for object recognition tasks. We hypothesize that its accuracy in object recognition can be improved due to the integration of both vision and tactile perception. 
We conducted experiments involving interactions between the sensor and 21 indentors of diverse 3D shapes to evaluate the ViTacTip's effectiveness in capturing detailed information for object recognition \cite{gomes2020geltip}. Examples of these interactions are illustrated in Fig. \ref{hybrid tactile sensing} (a). Through ViTacTip's real-time imaging, the localization and contour of each object are discernible. Concurrently, the distribution of markers on the sensor's skin changes, reflecting the shape of the contact.

Objects with shapes like `Grids', `Grates', `Curves', and `Dots' (see Fig. \ref{hybrid tactile sensing} (b)) present significant challenges to pure tactile perception. These shapes appear indistinguishable when relying solely on the deformation detected by the sensor's skin, due to the limited resolution of pin-like markers. This is where the ViTacTip's visual capabilities become critical. By providing a detailed visual representation of these complex shapes, the ViTacTip effectively overcomes the limitations of tactile-only perception, highlighting its multi-modal advantage in distinguishing objects that are difficult to differentiate through touch alone.

We initiated the experiments by collecting training data. This involved securely positioning the target objects, referred to as stimuli, on a stable surface.  The ViTacTip sensor was then applied to each stimulus at predetermined contact poses to ensure a standardized data collection process.
The center point position of each object's surface was defined as [0, 0, 0, 0], in terms of [$X$(mm), $Y$(mm), $Z$(mm), $\theta$($\degree$)]. For each object, 500 images were collected, resulting in a total of 10,500 images for 21 objects. The range of contact poses between the ViTacTip sensor and the object was defined as [5, 5, 1, 90] to [-5, -5, -1, -90]. To ensure precise deformation for accurate sensing, the sensor was progressively pressed down by 5mm at each contact point. Contact positions were systematically determined by evenly sampling within these predefined ranges, providing a uniform dataset representing the sensor's response to various points of contact.
Following data collection, we fine-tuned DenseNet121 \cite{iandola2014densenet} to process the images captured by ViTacTip. The embeddings extracted during this process were subsequently passed through a softmax function for classifying different objects.

\subsubsection{Results Analysis}

 In the object recognition task, ViTacTip achieved a nearly perfect accuracy at 99.91\%, which far exceeds the 88.03\% achieved by TacTip. As can be seen from Fig. \ref{hybrid tactile sensing} (a), the visual information within the multi-modality feature plays a significant role, as it clearly shows the contours of the object below, as well as the fine details of the contact surface. The spatial resolution and feature sensitivity of this pixel-level information surpass that of TacTip, which relies solely on sparse marker patterns. 
This advantage is particularly evident for objects with similar general shapes but differing fine details, such as the Cuboid (18) and the Hexagon (19). By integrating visual and tactile data, ViTacTip achieved superior recognition and differentiation of complex shapes, highlighting its potential as a powerful tool for automated object recognition tasks.

\subsection{Multi-Modality Perception for Hardness, Material, and Texture Recognition}
 \label{multi}
\subsubsection{Task Description}

The unique strength of the ViTacTip sensor lies in its ability to fuse multiple modalities. To thoroughly evaluate the effectiveness of this design, we conducted a comprehensive assessment of its multi-modality sensing capabilities.

For this evaluation, we fabricated five distinct elastomers using 3D printing technology. These elastomers were designed with varying hardness levels, measured on the Shore A scale, ranging from 17.5HA to 33.25HA. Each elastomer measured 30mm × 30mm × 10mm and was securely affixed to a rigid base to ensure stability during testing.
Additionally, a diverse set of fabric samples was used to evaluate the sensor's texture recognition capabilities, comprising 50 distinct textures.

During the data collection and evaluation process, each fabric sample was placed over the elastomers. This setup enabled the ViTacTip sensor to visually capture the fabric’s surface patterns while discerning the hardness of the underlying elastomer  through tactile sensing. By isolating these two modalities, the experimental design effectively assessed the sensor's visual and tactile performance independently.
To ensure comprehensive testing, all possible combinations of fabrics and elastomers were evaluated, resulting in a total of 250 combinations. This approach enabled the generation of a dataset that thoroughly assessed the ViTacTip sensor's ability to visually distinguish different textures and recognize various hardness levels via tactile perception.

The experimental setup and examples of the collected data are illustrated in Fig. \ref{hybrid tactile sensing} (c). These results demonstrate that the ViTacTip sensor captures not only tactile information but also visual features, such as the color and design of contact objects. The captured images facilitate a detailed analysis of the spatial arrangement and deformation of tactile elements, which is essential for precise tactile perception.
This approach significantly enhances the sensor's ability to interpret complex tactile data by providing visual context, thereby improving its utility in applications requiring accurate object identification and precise manipulation capabilities.

In this task, the pose range was defined within bounds of [3, 3, 0, 90] and [-3, -3, 0, -90] for [$X$(mm), $Y$(mm), $Z$(mm), $\theta$($\degree$)]. For each combination within these bounds, 100 images were collected, resulting in a total of 25,000 images across 250 combinations. The dataset was subsequently divided into training, validation, and test sets with a 5:3:2 ratio. These datasets were used to train and evaluate a multi-task learning model, as detailed below.

\begin{figure}[!htbp]
\captionsetup{font=footnotesize,labelsep=period}
	\centering
\includegraphics[width = 1\hsize]{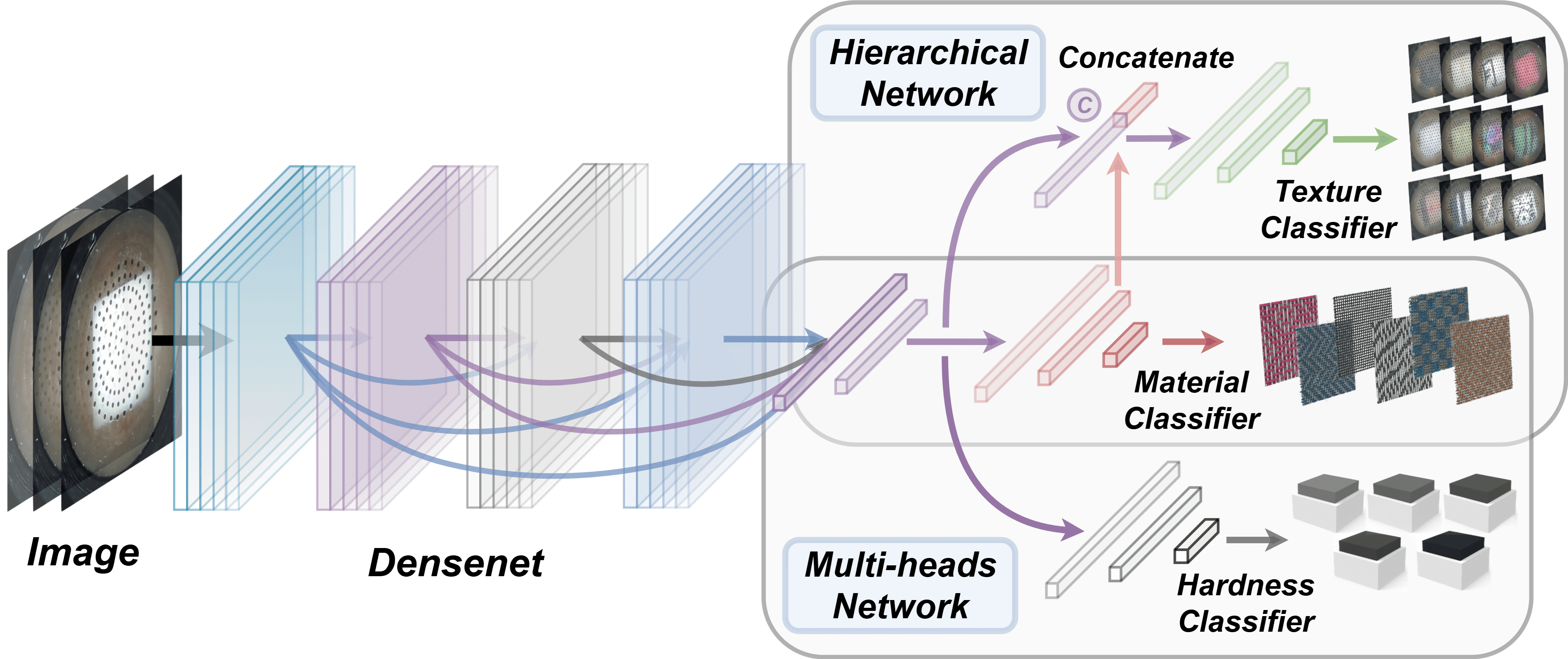}
	\caption{Neural network architecture of hierarchical multi-task learning for material, texture, and hardness recognition.}
 %\vspace{-0.2cm}
	\label{hierarchical}
\end{figure}

Our network architecture employs a hierarchical multi-head structure, where each head specializes in classifying one specific property: hardness, material, or texture. The network leverages the feature vector output from DenseNet \cite{huang2017densely}, which is distributed among the three sub-network heads. It is important to recognize the hierarchical relationship between material and texture. While different fabrics may be woven from the same material, they can exhibit a wide range of textures. Material properties often require a combination of tactile and visual perception to infer general features, such as the fibrous texture of wool, the smoothness of silk, and the roughness of hemp. In contrast, specific textures rely more heavily on visual information, including pattern geometry and color distribution, which are unique to each fiber type. This hierarchical relationship between material and texture properties is reflected in the design of the network. Details of the network structure and its implementation are illustrated in Fig. \ref{hierarchical}.

\subsubsection{Results Analysis}

After training, the network achieved recognition rates of 98.81\% for material identification and 97.78\% for texture identification. These high accuracy rates are particularly noteworthy considering the task's complexity: the sensor was required to differentiate among 50 distinct fabric patterns while these fabrics were placed over elastomers with varying hardness properties. The presence of the elastomers beneath the fabrics added an additional layer of complexity, potentially hindering the sensor's ability to accurately identify material and texture. Despite these challenges, the network demonstrated remarkable robustness and efficacy.

Moreover, the network successfully recognizes varying levels of hardness between elastomers, achieving an accuracy of 97.47\%. This result highlights the sensor's ability to discern subtle hardness variations through contact. 
The sensor's internal camera efficiently captures contact deformation, using markers to translate tactile information into measurable data via image processing. These results highlight the ViTacTip sensor's potential in applications such as cloth manipulation, where accurate identification of materials and textures is essential.

\begin{table}[ht]
\centering
\captionsetup{font=footnotesize,labelsep=period}
\caption{Hardware Benchmarking Result}
\label{tab:hardware_benchmarking}
\setlength\tabcolsep{4pt} % default value: 6pt
\begin{tabular}{lccccccccc}
\toprule
%\rowcolor{gray!20}
\textbf{Sensors} & \multicolumn{1}{c}{\textbf{Grating ↑}} & \multicolumn{4}{c}{\textbf{Pose Err. (mm)↓}} & \multicolumn{4}{c}{\textbf{Contact Pt. Err. (mm)↓}} \\
\cmidrule(lr){2-2} \cmidrule(lr){3-6} \cmidrule(lr){7-10}
& Acc & $X$ & $Z$ & $\theta$& Ave & $P_x$ & $P_y$ & $P_z$ & Ave \\
\midrule
Tactip    & 94.60\% & 0.16 & 0.11 & 0.47 & 0.25 & 0.56 & 0.57 & 0.10 & 0.41 \\
ViTac     & 99.72\% & 0.39 & 0.12 & 0.34 & 0.28 & 0.43 & 0.43 & 0.07 & 0.31 \\
ViTacTip  & \textbf{99.72\%} & \textbf{0.14} & \textbf{0.08} & \textbf{0.24} & \textbf{0.15} & \textbf{0.34} & \textbf{0.34} & \textbf{0.08} & \textbf{0.25} \\
\bottomrule
\end{tabular}
% \vspace{-0.1cm}
\end{table}

\begin{figure*}[!htbp]
\captionsetup{font=footnotesize,labelsep=period}
		\centering
		\begin{tabular}[b]{c}
\hspace{0cm}\includegraphics[width=1\textwidth,trim={0 0 0 0},clip]{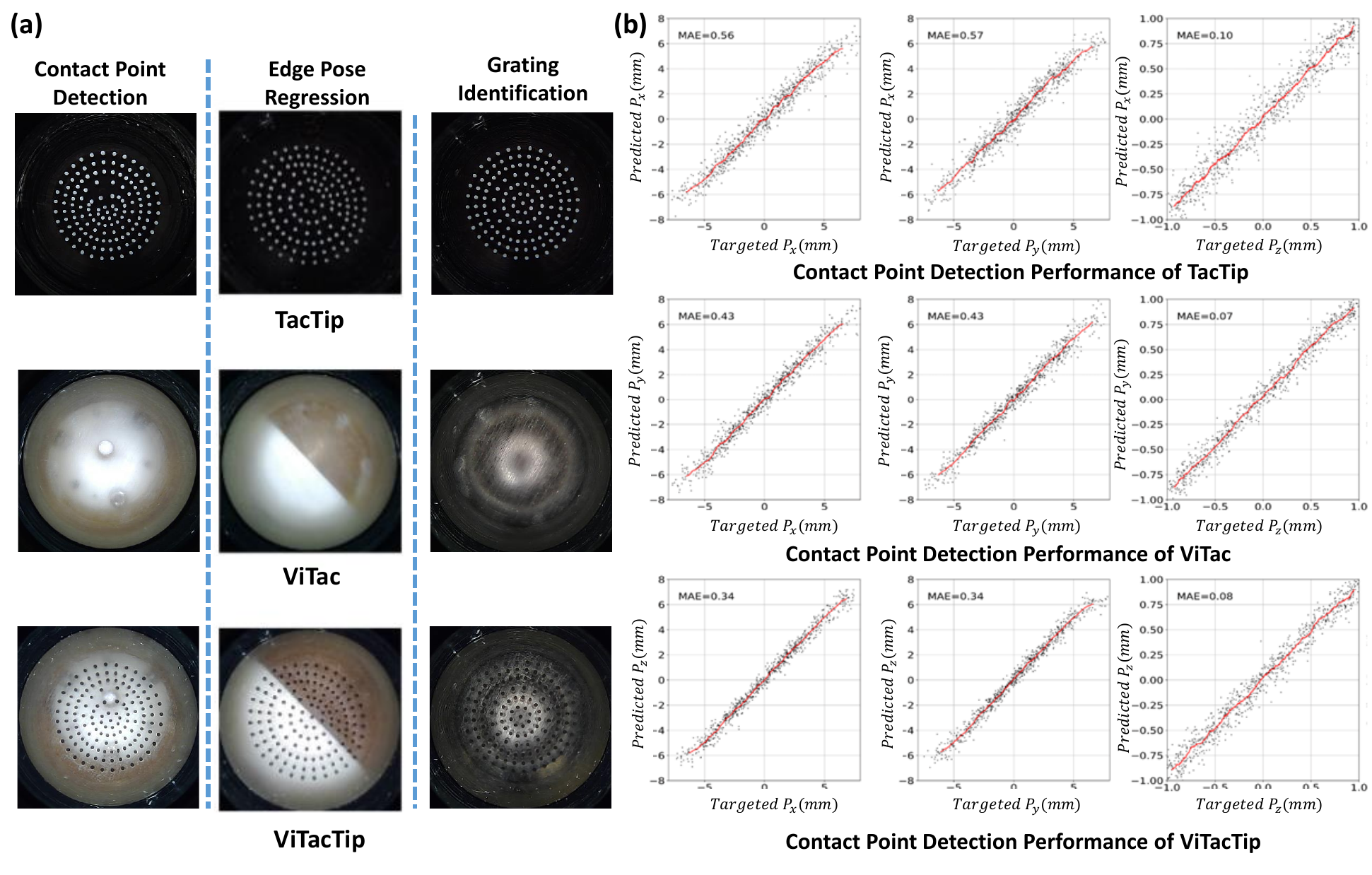} \\
		\end{tabular}
             \caption{ (a) Overview of the samples collected using TacTip, ViTac, and ViTacTip for hardware benchmarking experiments, including contact point detection, pose regression, and grating identification. (b) Comparative analysis of TacTip, ViTac, and ViTacTip based on MAE for the contact point detection task ($P_x$, $P_y$, $P_z$). }
		\label{contact-results}%
	% \vspace{-0.1cm}
\end{figure*}

\section{Hardware Benchmarking}
\label{hardware}

\subsection{Overview}
In this section, we present hardware benchmarking experiments to showcase ViTacTip's properties across three tasks: i) contact point detection, ii) pose regression, and iii) grating identification. 
For comparative analysis, we included two additional tactile sensors. The first sensor, referred to as \textbf{TacTip} \cite{lepora2021soft}, features pin-shaped markers embedded in an opaque (black) skin. To ensure a fair comparison, the TacTip was configured with the same marker distribution as the ViTacTip, consisting of seven concentric circles of markers (137 in total) and a skin thickness of 0.8 mm. This configuration balances contact sensitivity and wear resistance. The opaque skin of the TacTip effectively blocks out external light sources.
The second sensor, referred to as \textbf{ViTac}, uses only transparent skin and differs from the ViTacTip by excluding internal markers while maintaining the same skin setup, including thickness and material composition.

To ensure fair and consistent comparisons, we used DenseNet121 as the backbone architecture for all tasks and sensors. The comparative results provide valuable insights into the advantages of the ViTacTip and identify potential areas for improvement in practical applications. Samples collected from the TacTip, ViTac, and ViTacTip sensors during the hardware benchmarking experiments are shown in Fig.~\ref{contact-results}(a).

\subsection{Contact Point Detection}

This experiment focused on accurately predicting the relative spatial location of the stimulus' contact point on the sensor's skin, represented by the coordinates ($Px$, $Py$, $Pz$).
In a comparative analysis of performance metrics among TacTip, ViTac, and ViTacTip, ViTacTip demonstrates superior accuracy in detecting the contact point on the $x$ and $y$ axes, achieving the smallest mean errors of 0.34mm in both dimensions. In contrast, TacTip exhibits the largest contact point estimation errors across all three dimensions, with values of 0.56mm, 0.57mm, and 0.10mm, respectively.

The comparative results for contact point localization using the three sensors (TacTip, ViTac, and ViTacTip) are shown in Table \ref{tab:hardware_benchmarking} and Fig.~\ref{contact-results}(b). ViTacTip demonstrated exceptional precision in localizing contact points, even under challenging conditions such as varying shear forces and changes in lighting. This high performance is primarily attributed to the incorporation of additional visual information, which effectively mitigates positioning errors caused by shear movements and ensures reliable and consistent results.

\begin{figure*}[!htbp]
	\centering
 \captionsetup{font=footnotesize,labelsep=period}
	\includegraphics[width = 0.9\hsize]{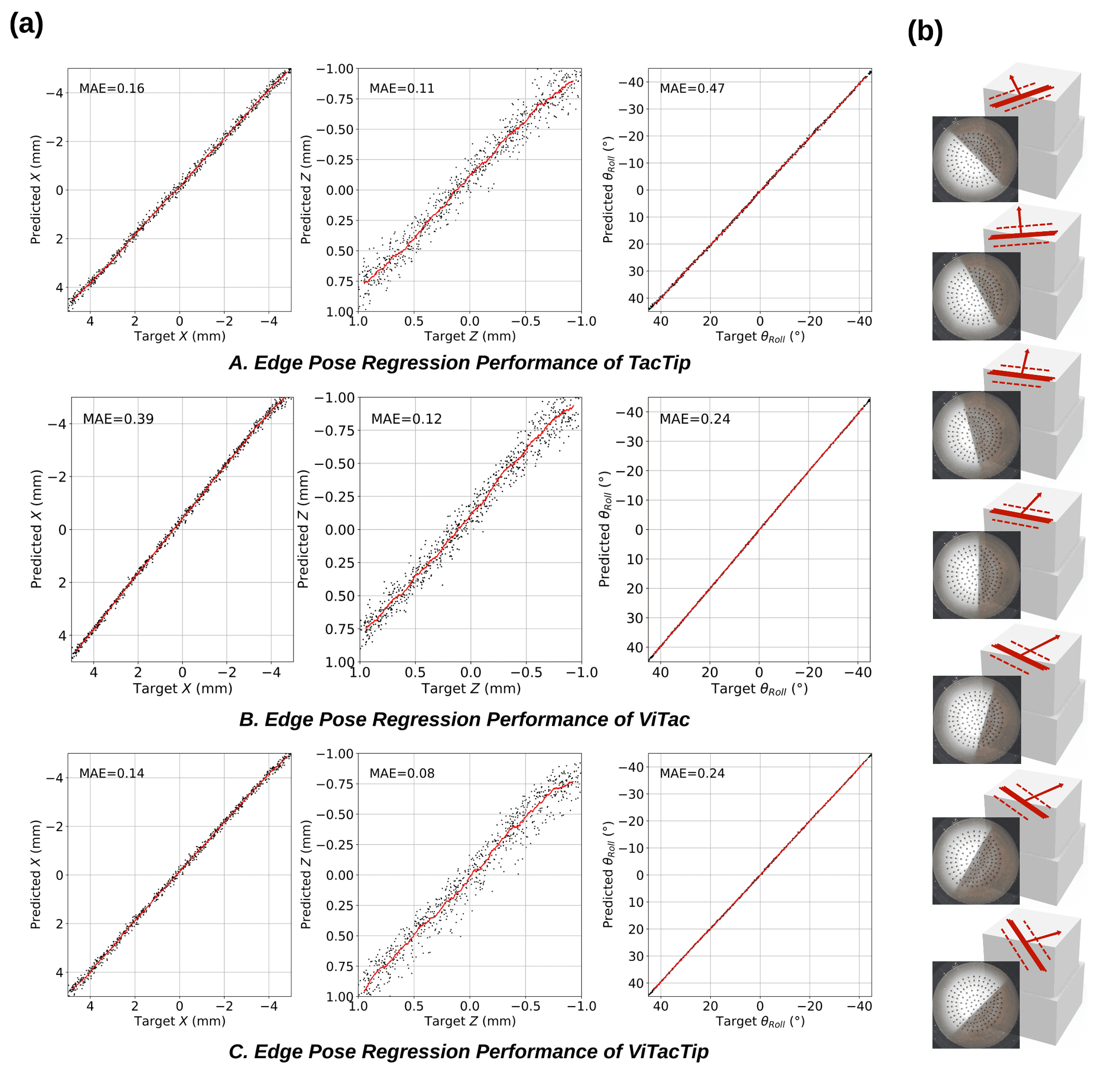}
	\caption{(a) Evaluation result of TacTip, ViTac, and ViTacTip on pose regression for horizontal distance $X$, press depth $Z$, rotation angle $\theta$. (b) Samples for the pose regression task using ViTacTip sensor.}
\vspace{-0.2cm}
	\label{pose error}
\end{figure*}

\subsection{Pose Regression}

To effectively evaluate the sensor's proprioception capabilities, pose regression tests were conducted \cite{lepora2021pose}. Previous studies have shown that the TacTip sensor is capable of pose regression \cite{lepora2021pose,9058673,lloyd2023pose}. Building on this foundation, we conducted a regression analysis using deep learning methodologies, focusing on predicting the sensor's pose relative to the boundary of a square stimulus.

The measured parameters included the horizontal distance ($X$) from the sensor's center to the boundary, the depth of press ($Z$) along the Z-axis, and the angle of rotation ($\theta$) along the Z-axis. We collected 3,000 images each for ViTac, TacTip, and ViTacTip, resulting in a total of 9,000 images. The dataset was randomly divided into training, validation, and test sets in an 8:1:1 ratio. To ensure a diverse range of pose values, $X$(mm), $Z$(mm), and $\theta$($\degree$), were varied within the ranges of [-5, 5], [-1, 1], and [-45, 45], respectively.

The comparative performance of the three sensors in pose regression is presented in Fig.~\ref{pose error}(a), while examples of different pose values represented by various perception images are visualized in Fig.~\ref{pose error}(b).
Compared to TacTip, ViTacTip demonstrates substantial improvements in reducing pose regression errors for $X$ (mm), $Z$ (mm), and $\theta$ ($\degree$), achieving reductions from 0.16 mm/0.11 mm/0.47$\degree$ to 0.14 mm/0.08 mm/0.24$\degree$. The most significant enhancement is observed in orientation estimation, with a 49\% improvement.

Although ViTac surpasses TacTip in $\theta$ estimation, its lack of a pin-like structure leads to the largest errors in $X$ and $Z$ estimations among the three sensors, reaching 0.39 mm and 0.12 mm, respectively. These results indicate that TacTip and ViTacTip, both incorporating pins and markers, are more effective in amplifying deformations caused by skin pressing against edges, enhancing pose regression accuracy. Overall, ViTacTip achieves the best performance.

\begin{figure*}[t]
	\centering
 \captionsetup{font=footnotesize,labelsep=period}
\includegraphics[width = 1\hsize]{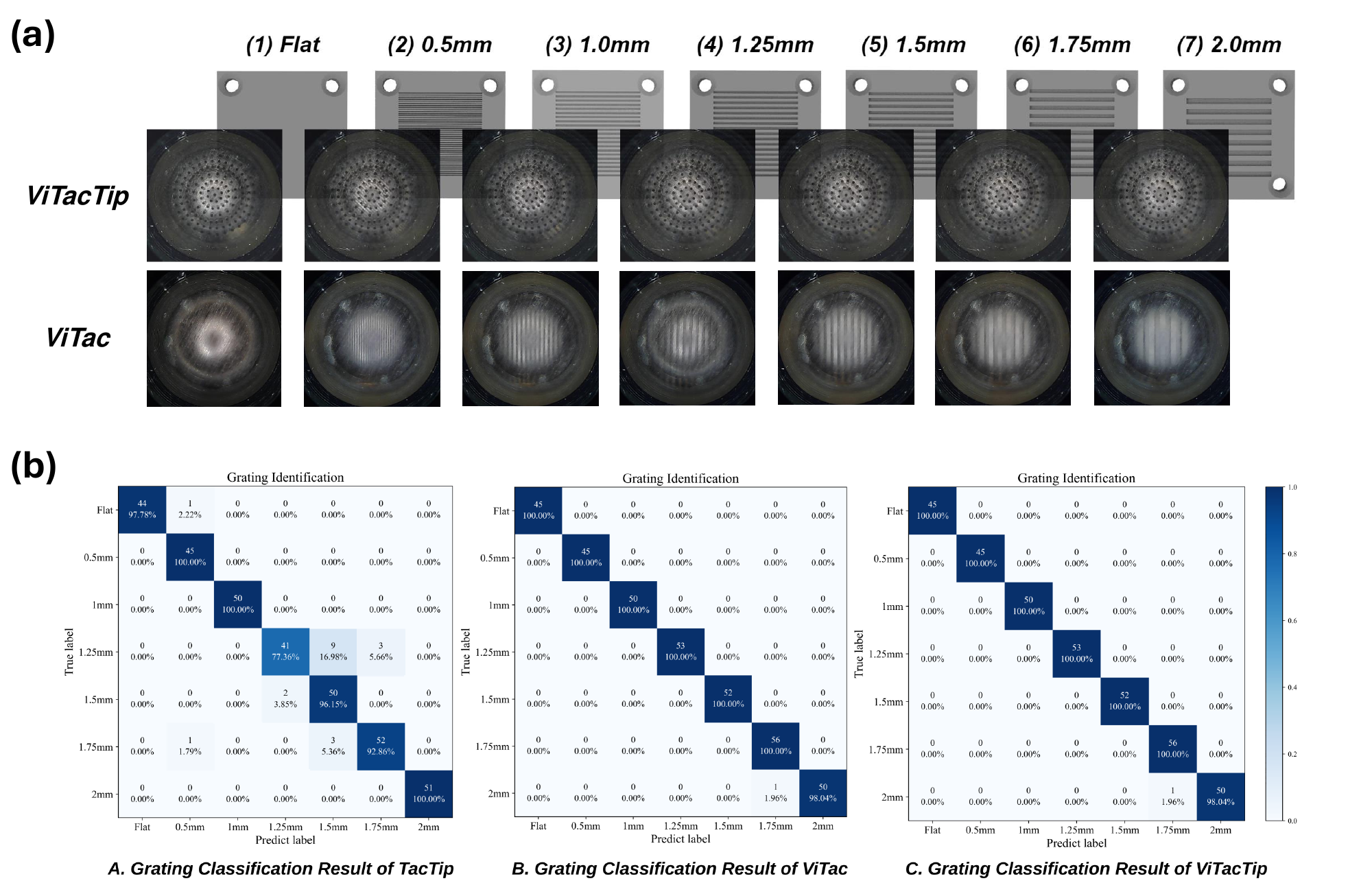}
	\caption{(a) Samples for the grating identification task and the perception images obtained using ViTacTip and ViTac. (b) Evaluation results of TacTip, ViTac, and ViTacTip on grating identification, presented as a confusion matrix.}
\vspace{-0.1cm}
	\label{grating evaluation}
\end{figure*}

\subsection{Grating Identification}

To estimate the spatial resolution of the ViTacTip, our study employed objects with various levels of detail as outlined in \cite{yuan2017gelsight}. 
A set of grating boards with different density specifications was applied to evaluate the spatial resolution of our proposed sensor. %~\ref{grating evaluation}
As shown in Fig.~\ref{grating evaluation} (a), we categorized the spatial resolution of these boards into three groups for tactile sensing: millimeters (0~mm, 1~mm, 2~mm), half-millimeters (0~mm, 0.5~mm, 1~mm, 1.5~mm, 2~mm), and quarter-millimeters (1~mm, 1.25~mm, 1.5~mm, 1.75~mm, 2~mm). In our analysis, we processed a substantial dataset of 3,500 data points, divided across the 7 categories of grating board specifications (0~mm/Flat, 0.5~mm, 1~mm, 1.25~mm, 1.5~mm, 1.75~mm, 2~mm). The data was split in a 7:2:1 ratio for training, validation, and testing.

Our research revealed that the TacTip sensor exhibited lower precision in detecting quarter-millimeter measurements. It struggled to accurately distinguish spacings of 1.25 mm, 1.5 mm, and 1.75 mm, resulting in a reduced test accuracy of 94.60\%. In contrast, both the ViTac and ViTacTip sensors demonstrated significantly better performance, achieving an impressive test accuracy of 99.72\%. This comparison underscores the superior sensitivity and precision of the ViTac and ViTacTip technologies in tasks requiring fine spatial resolution.

The improvement in spatial resolution achieved by ViTacTip, compared to TacTip, can be attributed to its integration of visual features. These visual capabilities enable ViTacTip to overcome certain limitations inherent to TacTip, such as constraints related to marker density and the conformability of the sensor's skin. By combining visual data with tactile sensing, ViTacTip significantly enhances its ability to discern fine spatial differences.
In addition to the results presented in Table \ref{tab:hardware_benchmarking}, the confusion matrices for the grating identification task performed by the three sensors (TacTip, ViTac, and ViTacTip) are shown in Fig.~\ref{grating evaluation} (b).

\section{GAN-Based Multimodal Interpretation}
\label{GAN}
\subsection{Overview}

The physical multi-modality fusion approach used in ViTacTip offers both advantages and disadvantages. For instance, while the marker layout in ViTacTip may not be dense enough to discern fine tactile/force-relevant information effectively, the additional visual modality compensates by providing richer details, such as color and fine texture-relevant features. However, the presence of black markers can introduce noise into the visual data, potentially distorting visual perception.
Additionally, the sensor's transparent skin, even with its internal light source, is still vulnerable to interference from external ambient lighting. 
The colorful background further complicates conventional tactile processing techniques, such as marker detection and tracking.
To address these challenges, we proposed two GAN-based solutions: one for marker removal and another for light interference mitigation.

\begin{figure*}[!htbp]
	\centering
\captionsetup{font=footnotesize,labelsep=period}
	\includegraphics[width = 0.95\hsize]{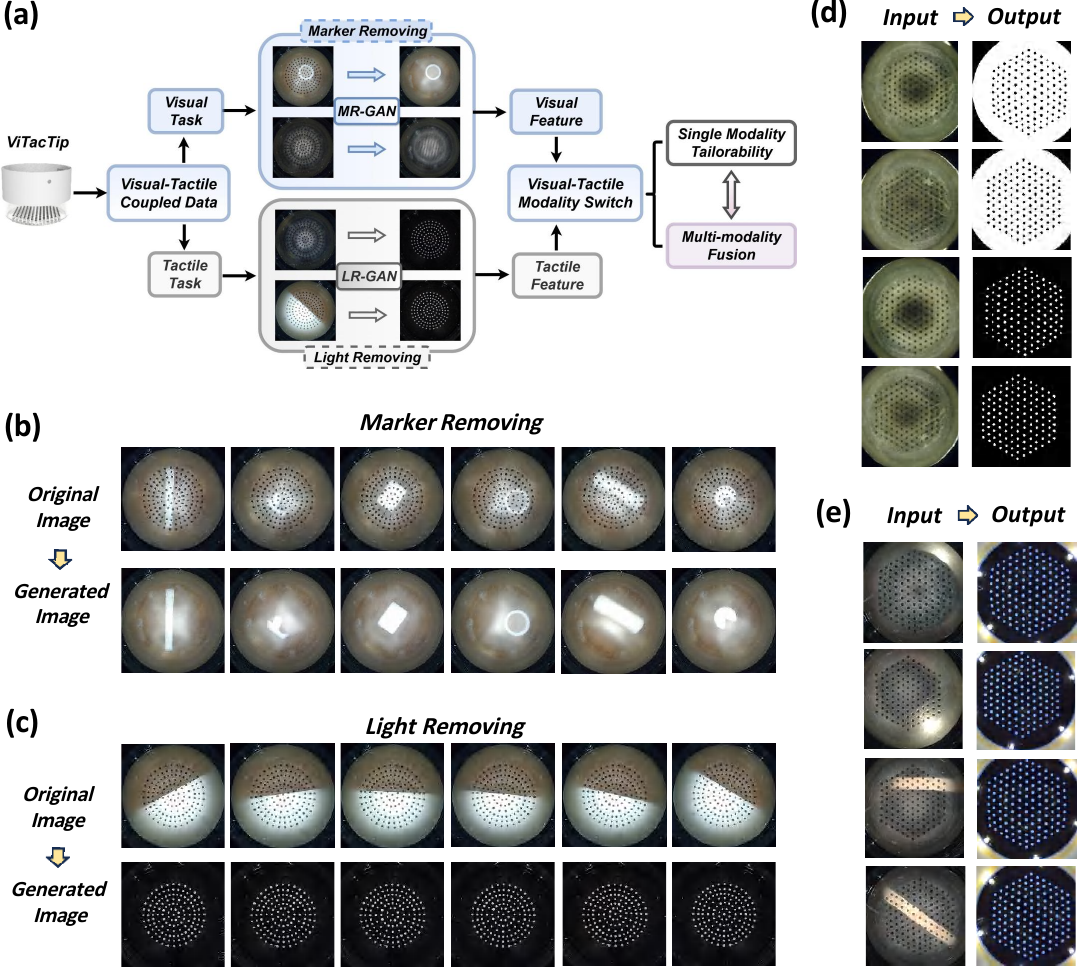}
	\caption{(a) Overview of the modality-switch framework: Two independent GAN models are trained on datasets from ViTac, ViTacTip, and TacTip to perform marker removal and light removal.
(b) Results of the MR-GAN modality switch for the object recognition task, converting ViTacTip-style images into ViTac-style images.
(c) Results of the LR-GAN modality switch for the pose regression task, converting ViTacTip-style images into TacTip-style images.
(d) Transformation of ViTacTip images into binary outputs isolating the tips, showcasing the versatility of the proposed GANs.
(e) Demonstration of consistent image quality in GAN-generated outputs, regardless of variations in ambient lighting conditions.}
 \vspace{-0.1cm}
	\label{Results-Gan}
\end{figure*}

\subsection{Method Description}

We collected three distinct datasets for our experiment using the ViTacTip, ViTac, and TacTip sensors. The TacTip sensor demonstrates minimal sensitivity to external light sources but lacks visual perception capabilities.
In contrast, the ViTac sensor, which is designed without internal pins or markers and consists solely of transparent skin, exhibits higher sensitivity to ambient light due to its simplified design. However, it compensates for this limitation by offering enhanced visual perception capabilities, allowing for improved detection of the objects it interacts with.

We trained the Marker Removal GAN (MR-GAN) to transform data from ViTacTip to ViTac. The output from MR-GAN replicates ViTac’s data, effectively eliminating occlusions caused by markers. Similarly, the Light Removal GAN (LR-GAN) processes ViTacTip data to emulate TacTip, mitigating the effects of ambient light interference. At the core of this approach is the separate extraction of visual and tactile information. This separation enhances the sensor’s adaptability, enabling it to be tailored for a wide range of applications.

Both MR-GAN and LR-GAN employ the Pix2Pix GAN architecture \cite{isola2017image} for perception image translation tasks.  Pix2Pix GAN conditions the generation of an output image on a corresponding input image, making it well-suited for modality switching among the ViTacTip, ViTac, and TacTip sensors.
The MR-GAN and LR-GAN networks comprise two main components: a Generator (G) and a Discriminator (D). The generator adopts a U-Net-style architecture, while the discriminator is implemented as a PatchGAN \cite{isola2017image}. The objective of the Pix2Pix GAN is formulated as a mini-max game between the generator and the discriminator.

\subsection{Results Analysis}
\subsubsection{Qualitative Results Analysis}
Fig. \ref{Results-Gan} (a) demonstrates that MR-GAN effectively removes all markers from ViTacTip data collected during object recognition and grating identification tasks, while preserving fine perception image features through visual perception. The resulting images closely resemble those obtained directly from the ViTac sensor. These results indicate that MR-GAN successfully eliminates markers and retains intricate texture details, significantly improving the clarity of tactile information, which is essential for enhancing accuracy in contact visualization during object recognition.
Additional examples of marker removal using MR-GAN for the object recognition task are presented in Fig. \ref{Results-Gan} (b).

Similarly, Fig. \ref{Results-Gan} (a) illustrates the processing of ViTacTip data for grating identification and pose regression tasks using LR-GAN. The transformed data successfully emulates the TacTip style by replacing the background and external lighting with a black skin and white markers. This transformation enables the application of marker-based tactile algorithms to ViTacTip, enhancing its robustness against ambient light interference and improving its functionality in varied lighting conditions.
More examples for the pose regression task, shown in Fig. \ref{Results-Gan} (c), demonstrate the successful conversion of ViTacTip data into TacTip-style data, featuring pure markers. This highlights the versatility and effectiveness of LR-GAN in adapting ViTacTip data for marker-based tactile processing.

\begin{figure*}[!htbp]
	\centering
\captionsetup{font=footnotesize,labelsep=period}
	\includegraphics[width = 0.9\hsize]{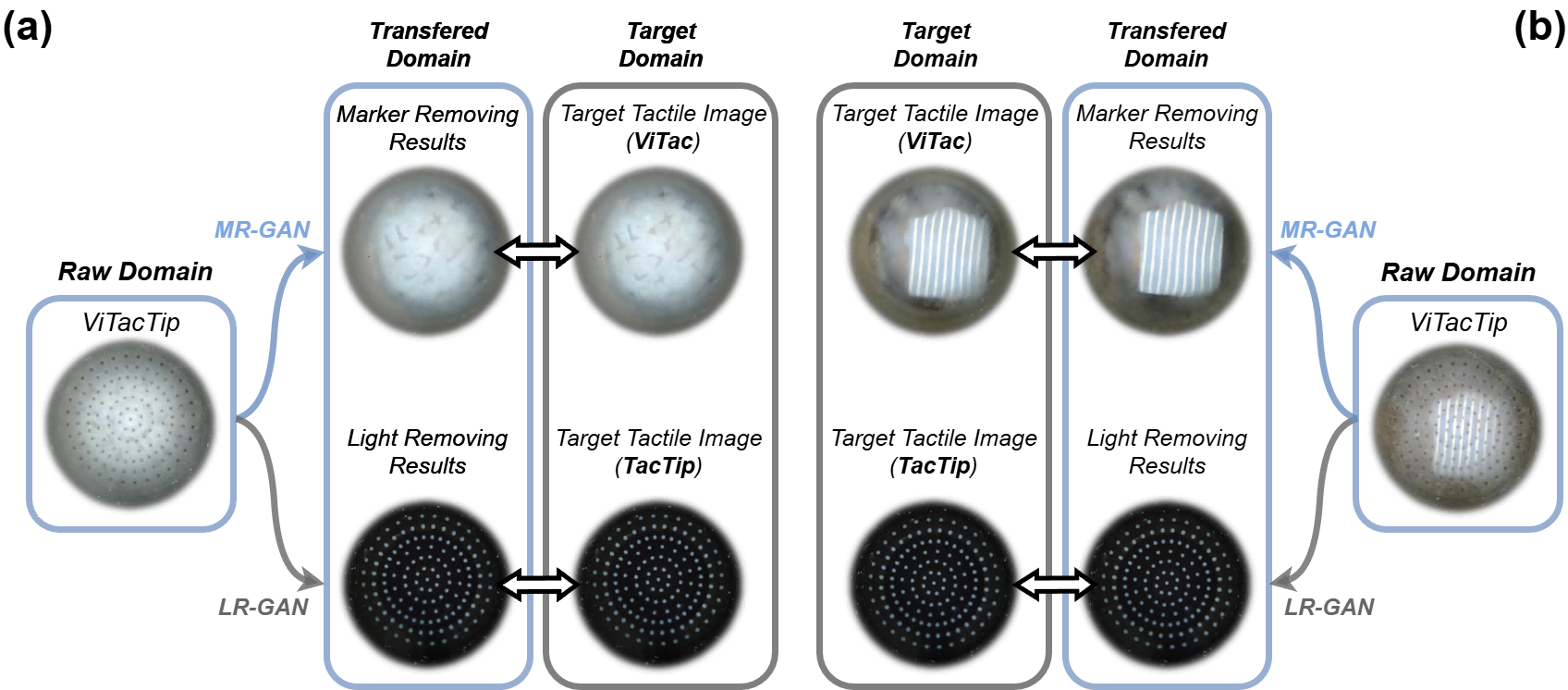}
	\caption{Examples of modality conversion when applying MR-GAN and LR-GAN to cube-like objects with complex textures: (a) irregularly shaped patterns and (b) striped cloth.}
% \vspace{-0.4cm}
	\label{Results-Gannew}
\end{figure*}

Additional results on modality switching and the generation of various data visualization formats are presented in Fig. \ref{Results-Gan} (d). As shown in Fig. \ref{Results-Gan} (e), the GAN-generated images maintain consistent quality even under varying ambient lighting conditions.
Fig. \ref{Results-Gannew} provides examples of modality conversion using MR-GAN and LR-GAN on cube-like objects with irregular-shaped patterns and striped cloth-like patterns. These results demonstrate the proposed method's applicability to complex scenarios, highlighting its versatility and robustness.

\subsubsection{Quantitative Results Analysis}

\begin{table}[!htbp]\centering
\renewcommand{\arraystretch}{1.2} % Reduced line spacing
\captionsetup{font=footnotesize,labelsep=period}
\caption{Quantitative Evaluation: Comparison of ground-truth data with LR-GAN and MR-GAN-based modality transfer results. `Texture Data' refers to data from cube-like objects with irregular-shaped patterns. }  
\label{gan similarity}
\begin{tabular}{lccc}
\toprule
\textbf{Scenarios} & \textbf{MSE}$\downarrow$ & \textbf{PSNR}$\uparrow$ & \textbf{SSIM}$\uparrow$  \\ \bottomrule
\textbf{MR Similarity on Grating Data}    & 0.0405    & 62.078    & 0.805  \\ 
\textbf{MR Similarity on Object Data}    & 0.0645    & 60.125    & 0.837  \\
\textbf{MR Similarity on Texture Data}      & 0.0105    & 70.534    & 0.932  \\ \midrule
\textbf{LR Similarity on Grating Data} & 0.0104    & 70.063    & 0.892  \\
\textbf{LR Similarity on Pose Data}    & 0.0429    & 62.663    & 0.708  \\ 
\textbf{LR Similarity on Texture Data}      & 0.0191    & 71.498    & 0.727  \\ \bottomrule
\end{tabular}
%\vspace{-0.2cm} % Adjusts spacing after the table
\end{table}

To evaluate the quality of the data generated by the GAN models, 40 images were randomly selected from each task for quantitative analysis using three metrics: MSE, PSNR (Peak Signal-to-Noise Ratio), and SSIM.
MSE quantifies the average squared difference between the pixel values of the ground-truth image in the target domain (referred to as the `target image') and the corresponding image generated by the GAN model. A lower MSE value indicates greater similarity and better alignment between the two images, as it reflects smaller pixel-wise differences.
PSNR, expressed in decibels (dB), measures the peak error between the target and generated images. Higher PSNR values indicate better image quality and greater similarity. Specifically, a PSNR value exceeding 40 dB reflects extremely high similarity, while values below 20 dB denote low similarity.
SSIM evaluates the structural similarity between the target and generated images by considering luminance, contrast, and structure. SSIM values range from 0 to 1, with values closer to 1 indicating higher similarity and better visual quality.

According to the results summarized in Table \ref{gan similarity}, all PSNR values exceed 40 dB, and the average SSIM scores are greater than 0.7. These results confirm that the GAN-generated outputs achieve a high degree of similarity to the ground truth images.  In summary, the proposed GAN models successfully executed modality switching, performing both marker-removing and light-removing tasks across different scenarios, including grating boards, various objects, and random poses.

\begin{table}[!htbp]\centering
\captionsetup{font=footnotesize,labelsep=period}
\renewcommand{\arraystretch}{1.2} % Reduced line spacing
\caption{Performance evaluation of GAN models under different lighting conditions when applying MR-GAN and LR-GAN on cube-like objects with complex textures (striped cloth-like pattern).}
\label{tab:gan_performance_light1}
\begin{tabular}{lccc}
\toprule
\textbf{Scenarios} & \textbf{MSE}$\downarrow$ & \textbf{PSNR}$\uparrow$ & \textbf{SSIM}$\uparrow$  \\ \bottomrule
\textbf{MR, Fixed Light Source}  & 0.0383 & 63.077 & 0.926 \\ 
\textbf{MR, Random Light Source} & 0.0799 & 59.217 & 0.837 \\
\midrule
\textbf{LR, Fixed Light Source}  & 0.0208 & 65.568 & 0.954 \\
\textbf{LR, Random Light Source} & 0.0208 & 65.569 & 0.954 \\
\bottomrule
\end{tabular}
\vspace{-0.2cm} % Adjusts spacing after the table
\end{table}

\subsection{Robustness Evaluation of Proposed Method}
To further assess the robustness and capabilities of the GAN models in complex environments involving challenging tasks, their performance was evaluated under conditions with both fixed and random light sources.
The LR-GAN model demonstrated remarkable consistency, maintaining high image quality across both lighting scenarios. It achieved an average PSNR of 65.57 and an SSIM of 0.954, with an MSE of approximately 0.0208.
In contrast, the MR-GAN model exhibited some variability in performance. Under fixed lighting conditions, it achieved moderately high image quality with a PSNR of 63.08 and an SSIM of 0.926. However, under random lighting conditions, its performance declined slightly, with the PSNR decreasing to 59.22 and the SSIM dropping to 0.837. The MSE also increased to 0.0799, indicating a higher error rate and greater sensitivity to lighting changes compared to the LR-GAN model. Table \ref{tab:gan_performance_light1} summarizes these findings.

Both GAN models significantly enhance the adaptability of the ViTacTip sensor. However, the LR-GAN model demonstrates superior robustness across varying environmental conditions.
In contrast, the MR-GAN model performs effectively under consistent lighting but shows a decline in performance under variable lighting. This indicates the need for further refinements to enhance its robustness and achieve adaptability comparable to the LR-GAN model in dynamic environments.

\section{Discussions and Future Work}
\label{Discuss-Future}

\begin{table*}[]
\centering
\captionsetup{font=footnotesize,labelsep=period}
\renewcommand{\arraystretch}{1.2} % Reduced line spacing
\caption{Comparative analysis with state-of-the-art multi-modality sensors}
\label{tab:Comparisons}
\begin{tabular}{llllll}
\toprule
\textbf{Sensor} & \textbf{Sensing Approach} & \textbf{Resolution} & \textbf{Shape} & \textbf{Modality Coupling} & \textbf{Modality Switch} \\ \bottomrule
\textbf{FingerVision\cite{yamaguchi2016combining}} & Vision \& Markers & - & 2D & High & None \\

\textbf{STS \cite{9423118}} & Vision \& Coating & Micron & 3D & High & None \\
\textbf{Finger-STS \cite{9832483}} & Vision \& Coating \& Markers & Micron & 3D & High & None \\
\textbf{Kazuhiro's \cite{7487126}} & Vision \& TIR & Micron & 2D & None & Compound-eye Camera (Visable/IR) \\

\textbf{SpecTac \cite{9812348}} & Vision \& UV Markers & - & 2D & None & UV LEDs (On/Off) \\

\textbf{UVtac \cite{kim2022uvtac}} & Vision \& Coating \& UV Markers & - & 2D & None & UV LEDs (On/Off) \\

\textbf{StereoTac \cite{10214622}} & Vision \& Coating & Sub-mm & 3D & None & LEDs (On/Off) \\

\textbf{StereoTac \cite{10214622}} & Vision \& Coating & Sub-mm & 3D & None & LEDs (On/Off) \\
\textbf{TIRgel \cite{10224334}} & Vision \& TIR & Micron & 2D & None & LEDs (On/Off) \& Focus (Near/Far) \\
\textbf{VisTac \cite{10242327}} & Vision \& Coating & Micron & 2D & None & LEDs (On/Off)  \\
\textbf{ViTacTip (Ours)} & \textbf{Vision \& Markers} & \textbf{Micron} & \textbf{3D} & \textbf{Decoupled by GAN} & \textbf{MR-GAN \& LR-GAN} \\ \bottomrule
\end{tabular}
%\vspace{-0.3cm}
\end{table*}

\subsection{Discussions}

\subsubsection{Comparison with Single-Modality VBTS}
GelSight-type sensors are renowned for their high spatial resolution in tactile perception. Unlike the mold-cast elastomer skin of GelSight sensors, the ViTacTip features a 3D-printed skin \cite{lepora2022digitac}, enabling customizable shapes and dimensions. Its ultra-soft, gel-filled construction further enhances conformance to curved surfaces, making it ideal for tasks demanding precise force sensing, a large dynamic range, and safe object contact. 

Furthermore, the ViTacTip sensor also benefits from the easy modification of its pin density or shape through 3D printing, which can improve tactile perception capability. However, increasing the pin density may impact visual perception capabilities, as the pins can obstruct the detection of fine geometric details on targeted objects.
To address this, we need to find an optimal balance between tactile and visual perception capabilities through careful design of the markers and improvements in the generalizability of GAN-based modality switch approach.

Additionally, while maintaining the same volume, weight, and robustness as TacTip, the integration of multi-modal perception marks a significant advancement. This positions ViTacTip as the sensor of choice for applications demanding both detailed tactile information and visual perception of fine features.
The incorporation of multi-modal sensing elevates environmental interaction and object recognition by enabling the simultaneous detection of shape, material, and texture, rather than relying solely on touch-based data.

\subsubsection{Comparison with Multi-Modality VBTS}
Here, we conduct a comparative analysis of state-of-the-art multi-modality sensors developed between 2016 and 2023, as outlined in Table \ref{tab:Comparisons}. This analysis focuses on critical aspects of tactile sensor design, including the sensing approach, spatial resolution, surface shape, modality coupling, and the capability to switch between modalities.

Upon examination, ViTacTip can capture fine details of both tactile and visual in the captured frames, delivering rich data for subsequent analysis. Firstly, ViTacTip serves as a proficient proximity sensor, capable of perceiving precise visual information through its see-through-skin mechanism. This mechanism, coupled with high spatial resolution, is important for applications such as object and texture recognition. Additionally, the pin-shaped markers in ViTacTip exhibit high sensitivity to dynamic tactile events, effectively monitoring touch deformations caused by normal or shear forces. This feature is especially valuable when the sensor’s skin conforms to or moves along a touched surface. In contrast, sensors like STS \cite{9423118} and TIRgel \cite{zhang2023tirgel} lack force-sensing capabilities. ViTacTip, however, offers a more comprehensive suite of functionalities, including force regression, shape recognition, and proximity sensing.
In addition, unlike Kazuhiro's sensor \cite{7487126}, whose contact surface is flat, ViTacTip's pin-like markers can be easily printed on 3D curved surfaces, which can enhance its sensing capability in complex scenarios. 

Unlike other multimodal VBTSs \cite{yamaguchi2016combining, 9423118} that lack a modality-switching function, ViTacTip enables independent access to perception data from different modality domains.  This design effectively mitigates the challenges associated with modality coupling, enabling seamless extraction and utilization of individual modalities as required.  Through its GAN-based image modality translation, ViTacTip operates efficiently across multiple modes, including multimodal fusion, visual perception, and tactile perception.

Most other multimodal VBTSs, such as SpecTac \cite{9812348}, UVtac \cite{kim2022uvtac}, and TIRgel \cite{zhang2023tirgel}, depend on state-switching capabilities of external hardware, such as UV LEDs or camera focus adjustments. 
Hardware-based multimodal systems often face the challenge of missing significant information during modality switching gaps, particularly in dynamic scenarios. ViTacTip addresses this limitation by adopting a software-based approach, eliminating the redundancy and complexity associated with hardware design for multimodal sensors. By bypassing hardware dependencies, ViTacTip minimizes delays, significantly enhancing its efficiency and robustness in environments where high response speed is critical.
Notably, the GAN-based approach employed in ViTacTip not only optimizes its performance but also offers adaptability for integration with other types of multimodal sensors, thereby broadening its potential applications and versatility across diverse fields.

In summary, the ViTacTip sensor stands out with its unique blend of hardware functionalities and software innovations, surpassing traditional single-modality VBTSs and existing multimodal VBTSs through the pioneering use of GANs for the seamless modality switch. The integration of MR-GAN and LR-GAN ensures smooth transitions between modalities. In our proposed GAN-based modality switch framework, the MR-GAN enables ViTacTip to capture fine visual details without dot-like noise caused by the pin-shaped markers. Meanwhile, the LR-GAN improves ViTacTip's adaptability to diverse lighting conditions and environmental settings, demonstrating its performance in real-world applications where other multimodal VBTSs may struggle.

\subsection{Future Work}

We plan to evaluate the sensor's effectiveness in scenarios involving multiple contact points. Additionally, we will conduct comparative analyses with various types of VBTSs, focusing on aspects such as resolution, sensitivity, and adaptability to diverse environmental conditions. Comparing ViTacTip with established GelSight-type sensors is expected to provide a more comprehensive understanding of its performance. However, it is important to note that existing GelSight-type sensors involve numerous parameters that may influence their softness. For example, while the commercial DIGIT sensor is a GelSight-type sensor, it is not suitable for a fair comparison in this study due to its significantly smaller sensing area. In future research, we aim to develop standardized approaches for benchmarking tactile sensor hardware.

Furthermore, we will explore the integration of ViTacTip into various types of robotic hands. Its shape can be easily customized for seamless incorporation into the fingertips or palms of robotic hands. The multi-modal sensing capabilities of ViTacTip have the potential to significantly enhance robotic dexterity, making it a valuable tool for complex manipulation tasks. Additionally, ViTacTip can drive advancements in multimodal robot learning. This integration is particularly advantageous for applications requiring delicate handling, such as electronics manufacturing assembly lines, as well as laboratory and domestic settings where contact-rich manipulation is crucial.

\section{Conclusions}
\label{Conclusions}
In this study, we introduce ViTacTip, a novel sensor with multi-modal capabilities comprising two principal modalities (vision and tactile) and two derived modalities (proximity and force). Hardware benchmarking of ViTacTip across tasks such as grating identification, pose regression, and contact point detection demonstrates its versatility in a wide range of applications.

ViTacTip’s performance was benchmarked against the TacTip and ViTac sensors. In these comparisons, specific features such as the see-through-skin mechanism and biomimetic pins were intentionally omitted to align its design with those of TacTip and ViTac. This approach serves as an ablation study, allowing for the assessment of the impact of these features on overall performance.
Furthermore, a hierarchical multi-task learning framework highlights ViTacTip’s advanced capabilities, enabling it to simultaneously identify hardness, materials, and textures. These multi-modal capabilities extend beyond the limitations of TacTip and ViTac, which are confined to single-modality perception.

The integration of MR-GAN and LR-GAN models with ViTacTip’s sensor data represents a significant advancement. These GAN models empower the sensor to handle diverse tactile sensing tasks, including fine feature recognition and dynamic force estimation. By transforming ViTacTip’s data to emulate the output styles of TacTip and ViTac using LR-GAN and MR-GAN, respectively, the sensor achieves enhanced versatility and operational effectiveness in various scenarios.

\section*{Acknowledgment}

The authors would like to thank Andrew Stinchcombe and Tom Barnes for their help in sensor design and fabrication.

\bibliographystyle{IEEEtran}

\bibliography{IEEEabrv,ref}

\end{document}